\documentclass{article}

\usepackage{PRIMEarxiv}

\usepackage[utf8]{inputenc} 
\usepackage[T1]{fontenc}    
\usepackage{hyperref}       
\usepackage{url}            
\usepackage{booktabs}       
\usepackage{amsfonts}       
\usepackage{nicefrac}       
\usepackage{microtype}      
\usepackage{lipsum}
\usepackage{fancyhdr}       
\usepackage{graphicx}       
\graphicspath{{media/}}     
\usepackage{url}

\usepackage{caption}
\usepackage{subcaption}
\usepackage{bm}
\usepackage{color}
\usepackage{xcolor}
\usepackage{soul}
\usepackage{multirow}
\usepackage{amsmath}

\def\pari{\bm{\mu}}
\def\qoi{\bm{q}}
\def\dpt{\bm{u}}
\def\lfdp{\bar{\bm{u}}}
\def\hfdp{\bm{u}}
\def\lfdpi{\bar{\bm{u}}^{i}}
\def\hfdpi{\bm{u}^{i}}
\def\lfdpj{\bar{\bm{u}}^{j}}
\def\hfdpj{\bm{u}^{j}}
\def\mfdpi{\bm{w}^{i}}
\def\Nl{\bar{N}}
\def\Nh{N}

\title{A few-shot graph Laplacian-based approach for improving the accuracy of low-fidelity data}

\author{
  Orazio Pinti\\
  University of Southern California \\
  Los Angeles, CA\\
  \texttt{pinti@usc.edu} \\
  \And
  Assad A. Oberai \\
  University of Southern California \\
  Los Angeles, CA\\
  \texttt{aoberai@usc.edu} \\
}

\begin{document}
\maketitle

\begin{abstract}
Low-fidelity data is typically inexpensive to generate but inaccurate. On the other hand, high-fidelity data is accurate but expensive to obtain. Multi-fidelity methods use a small set of high-fidelity data to enhance the accuracy of a large set of low-fidelity data.
In the approach described in this paper, this is accomplished by constructing a graph Laplacian using the low-fidelity data and computing its low-lying spectrum. This spectrum is then used to cluster the data and identify points that are closest to the centroids of the clusters. High-fidelity data is then acquired for these key points. Thereafter, a transformation that maps every low-fidelity data point to its bi-fidelity counterpart is determined by minimizing the discrepancy between the bi- and high-fidelity data at the key points, and to preserve the underlying structure of the low-fidelity data distribution. The latter objective is achieved by relying, once again, on the spectral properties of the graph Laplacian. This method is applied to a problem in solid mechanics and another in aerodynamics. In both cases, this methods uses a small fraction of high-fidelity data to significantly improve the accuracy of a large set of low-fidelity data. 
\end{abstract}


\section{Introduction}
Multi-fidelity methods have been widely used in different areas of science and engineering, including optimization  \cite{Goel2007,gramacy2009adaptive}, uncertainty quantification \cite{Allaire_2014}, uncertainty propagation \cite{Ng2014,Peherstorfer2016}, and statistical inference \cite{Kaipio2007} (see Fernández-Godino et al. \cite{Godino_2019} and Peherstofer et al. \cite{Peherstorfer2016} for two comprehensive reviews).
The fundamental idea behind these methods is to combine a large amount of low-fidelity data, which is relatively cheap to compute or measure but may contain errors, with a smaller but accurate set of high-fidelity data. 
The model used to generate the low-fidelity data is generally inexpensive, as part of its accuracy is traded in favor of a lower computational burden. 
There are several ways to accomplish this, such as implementing simpler physics, using coarser space/time discretization, or applying a reduced-order modeling technique like proper orthogonal decomposition (POD).
On the other hand, the model used to generate high-fidelity data is accurate but expensive to deploy. The objective of multi-fidelity modeling is to obtain a representation of the system that retains most of the accuracy of the high-fidelity model at a much lower cost. 

In a typical multi-fidelity framework, the first step is to employ the low-fidelity model to investigate the input space and obtain an approximate distribution of the response of the system. Then, a limited number of high-fidelity data points are computed or measured. 
Finally, techniques that learn the response from the low-fidelity data, and correct it using the high-fidelity data are applied. Along these lines, co-kriging methods have been extensively investigated \cite{kennedy2000predicting, forrester2007multi, perdikaris2015multi, park2017remarks}. In these methods, a Gaussian process is used as a prior for both the low-fidelity response and a discrepancy function, which models the difference between the low- and high-fidelity data. Then, a Bayesian approach is used to update the posterior distribution in the light of the low- and high-fidelity observations.

Other methods make use of radial basis functions (RBFs) to model the low-fidelity response. Specifically, the low-fidelity surrogate is written as an expansion in terms of a set of RBFs, and the coefficients are determined by interpolating the available low-fidelity data. The multi-fidelity approximation is then obtained in different ways. These include determining a scaling factor and a discrepancy function, which can be modeled using a kriging surrogate \cite{park2017remarks}, or another expansion in terms of RBFs \cite{durantin2017multifidelity, song2019radial}. In some cases the multi-fidelity surrogate is constructed by mapping the low-fidelity response directly to the high-fidelity response \cite{zhou2017variable}.

More recently, deep neural networks have also been used to fit low-fidelity data and learn the complex map between the input and output vectors in the low-fidelity model. Then, the relatively small amount of high-fidelity data is used in combination with techniques such as transfer learning \cite{chakraborty2021transfer, de2020transfer}, embedding the knowledge of a physical law through a physics-informed neural loss function \cite{PENWARDEN2022110844, MENG2020109020, meng2021multi, RAISSI2019686}, or, in the case of multiple levels of fidelity, concatenating neural networks together \cite{li2020multi}. An approach that involves training a physics-constrained generative model, conditioned on the low-fidelity snapshots, to produce solutions that are higher-fidelity and higher-resolution has also been proposed \cite{Geneva2020}.

Another class of methods, that are applicable when the response consists of a vector of large dimension, perform order-reduction using the low-fidelity data, and then inject accuracy using the high-fidelity data in a reduced-dimensional latent space. 
For example, this can be done by computing the low- and high-fidelity POD manifolds, aligning them with each other, and replace the low-fidelity POD modes used to represent the response with their high-fidelity counterparts \cite{perron2020development}. 
This can also be accomplished by solving a subset selection problem to construct a surrogate model for the low-fidelity response in terms of a few, important snapshots \cite{narayan2014stochastic}.
The high-fidelity model is then deployed to generate the high-fidelity version of the important snapshots and replace them in the expansion for the system response, yielding a multi-fidelity model that has found applications in tasks like topology optimization \cite{keshavarzzadeh2019parametric} and modeling interactional aerodynamics in rotorcrafts \cite{pinti2022multi}. 

The approach developed and applied in this manuscript also follows the main steps outlined above. However, the method used to learn the distribution of the low-fidelity data and to select the points at which to compute the high-fidelity data relies on the spectral properties of the graph Laplacian. In particular, the low-fidelity data points are treated as nodes of a weighted graph where the adjacency matrix is evaluated using a Gaussian kernel and the Euclidean distance in the normalized data coordinates. Thereafter, the graph Laplacian is constructed and its low-lying eigenspectrum is computed. These eigenfunctions are used to cluster the low-fidelity data \cite{chung1997spectral}, driving the selection of which high-fidelity simulations are needed. 
Once the high-fidelity data is computed (or measured), the eigenfunctions of the graph Laplacian are used once again to determine a transformation of the low-fidelity data to their bi-fidelity counterparts. This ensures that the bi-fidelity data points inherit the structure of low-fidelity data, which is embedded in the eigenfunctions.

The approach described in this manuscript is applicable to problems where the response of the system is described by $d = O(1-10^2)$ quantities of interest. A typical example is an external flow over a body where the quantities of interest include aerodynamic forces, moments, and the location of the points at which the flow separates or transitions. When constructing a bi-fidelity model for a problem like this, our approach explicitly looks for the structure in the distribution of the output data, and utilizes this in constructing the bi-fidelity response. This is in contrast to most kriging and RBF-based methods described above, which would construct distinct models for each of these quantities. 
This approach also finds strong connections with recent works in semi-supervised classification algorithms on graphs \cite{bertozzi2018uncertainty, slepcev2019analysis, belkin2004regularization, bertozzi2021posterior, dunlop2020large}, and theoretical results on consistency of graph-based methods in the limit of infinite data that provide a mathematical foundation \cite{hoffmann2020consistency, trillos2018variational}. 

The remainder of the manuscript is organized as follows. In Section \ref{sec:background}, we summarize the necessary background in graphs and spectral clustering. In Section \ref{bi-fi-section}, we describe the proposed bi-fidelity approach. In Section \ref{sec:theory}, we summarize some important theoretical results for the proposed method. In Section \ref{num-experiment}, we test our method on sample problems. Finally, in Section \ref{conc} we present the conclusions of our work. 

\section{Background} \label{sec:background}
A complete, weighted graph is a pair $G=(V,\,\bm{W})$, where $V=\{\dpt^1,\,\dots,\,\dpt^N\}$ is a set of vertices (or nodes) in $\mathbb{R}^D$, and $\bm{W}=[W_{ij}]$ is an adjacency matrix. We consider adjacency matrices of the type
\begin{eqnarray}
W_{ij} = d (|| \dpt^i - \dpt^j || ), \label{eq:adj}
\end{eqnarray}
where $d(\cdot)$ is a monotonically decreasing function and $||\cdot||$ is the $l_2$ norm. We define the degree matrix $\bm{D}$ to be a diagonal matrix with 
\begin{eqnarray}
D_{ii} = \sum_{j = 1}^{N} W_{ij},
\label{eq:degree_mat}
\end{eqnarray}
and a family of graph Laplacians,
\begin{eqnarray}
\bm{L} = \bm{D}^{-p} (\bm{D} - \bm{W}) \bm{D}^{-q}.
\label{eq:GL}
\end{eqnarray}
Different choices of $p$ and $q$ result in different normalizations of the graph Laplacian \cite{belkin2006convergence, COIFMAN20065, trillos2018error}.  
The graph Laplacian can be used to perform spectral clustering, which amounts to finding an optimal partition of the graph using the spectral properties of $\bm{L}$ \cite{guattery2000graph, von2007tutorial}. The eigenfunctions of $\bm{L}$ form the set of orthonormal functions from the nodes of the graph to the real numbers $\bm{\phi}^{(m)}:\dpt \rightarrow \mathbb{R}$ that solve the eigenvalue problem
\begin{eqnarray}
\bm{L}\bm{\phi}^{(m)} = \lambda_m \bm{\phi}^{(m)},
\label{eq:eigenprob}
\end{eqnarray}
with $\bm{\phi}^{(m)} = [\phi^{(m)}_1,\,\dots,\,\phi^{(m)}_N]^T$ and $\phi^{(m)}_i=\phi^{(m)}(\dpt^i)$.
For the un-normalized graph Laplacian ($p=q=0$), they satisfy the following property,

\begin{align}
\bm{\phi}^{(m)} \cdot \bm{L} \bm{\phi}^{(m)} &= \lambda_m \\
&= \frac{1}{2} \sum_{i,\,j} W_{ij} \left(\phi^{(m)}_i - \phi^{(m)}_j \right) ^2. 
\label{eq:quad_form}
\end{align}

This implies that eigenfunctions with small eigenvalues provide a mapping of the graph to a line that promotes the clustering of vertices that are strongly connected. Note that the eigenvalue problem (\ref{eq:eigenprob}) admits the trivial solution that maps all vertices to a point, e.g. $\bm{\phi}^{(1)} = \frac{1}{\sqrt{N}}[1,\,\dots,\,1]^T$, and has a zero eigenvalue.  
The eigenfunction corresponding to the smallest non-zero eigenvalue, also called Fiedler vector, represents the non-trivial solution to the problem of embedding the graph onto a line so that connected vertices stay as close as possible \cite{belkin2001laplacian}. 
Similarly, the eigenfunctions corresponding to the $k$ lowest non-zero eigenvalues, $[\bm{\phi}^{(2)},\,\dots,\,\bm{\phi}^{(k+1)}]$, represent the optimal embedding of the graph into $\mathbb{R}^k$, where the coordinates of a vertex $\dpt^i$ are given by $\bm{\xi}_i = [\phi_i^{(2)},\dots,\phi_i^{(k+1)}]$.

\section{Bi-fidelity method}\label{bi-fi-section}
Given a parametric physical problem, suppose that it is possible to generate low- and high-fidelity solutions.
The process for generating low-fidelity data is inexpensive but possibly affected by error, whereas the high-fidelity process is accurate but resource intensive. We denote by $\pari \in \mathbb{R}^P$ the vector of input parameters, and by $\bar{\qoi} (\pari) \in \mathbb{R}^Q$ and $\qoi (\pari) \in \mathbb{R}^Q $ the low- and high-fidelity solutions, respectively. They represent a set of $Q$ output quantities of interest of the problem.

Low- and high-fidelity data vectors are built from components of input parameters and output quantities, and are denoted by $\lfdp \in \mathbb{R}^D$ and $\dpt \in \mathbb{R}^D$, respectively, where $1 < D \le Q+P $. That is, $\lfdp = \bm{R} [\pari, \, \bar{\qoi}]$ and $\dpt = \bm{R} [\pari, \, \qoi]$, where $\bm{R}$ is a restriction operator that extracts the appropriate components of the input parameters and output quantities of interest. The choice of $\bm{R}$ is problem dependent and is described in Appendix \ref{appx_dataspace}. Two obvious choices are $ \bm{R} [\pari, \, \qoi]  = [\pari, \, \qoi (\pari) ] $ and $\bm{R} [\pari, \, \qoi] =  \qoi(\pari)$. That is, the data comprises of all input and output components, or just the output components. 

Our goal is to combine a dense set of low-fidelity data points with a few, select high-fidelity points to construct a dense set of bi-fidelity points that inherit accuracy from the high-fidelity process. 
To accomplish this we (1) generate a dense collection of low-fidelity data points, (2) identify key input parameter values at which to query the more expensive high-fidelity process, and (3) combine the low- and high-fidelity data to construct a bi-fidelity model. Below, we describe each step in detail. 

\paragraph{Step 1 -- Construction of the low-fidelity graph.} We begin by sampling $\bar{N} \gg 1$ points in the parameter space from a simple prescribed probability density to generate the set $\bar{S}=\{\pari^i\}_{i=1}^{\bar{N}}$. Then, we generate the low-fidelity data points $\lfdpi = \lfdp (\pari^i), i = 1, \dots, \bar{N}$, and collect them in the set $\bar{\mathcal{D}}=\left\{\lfdpi\right\} _{i=1}^{\bar{N}}$. 
For uniformity, we scale them so that each component lies within $[-1,\, 1]$. We apply the same scaling factors to the high-fidelity data collected in Step 2.

We treat each data point as the vertex of an undirected, complete, weighted graph, and exploit the useful properties of the associated matrices. Hence, we construct a graph with $\lfdpi$ as vertices, and with weights given by the the entries of the adjacency matrix (\ref{eq:adj}) where $d(\cdot)$ is chosen to be a Gaussian kernel,
\begin{equation}
d(r) \equiv \exp(-r^2/ \sigma^2).
\label{eq:distance}
\end{equation}
$\sigma$ is characteristic scale, which can be treated as a hyperparameter or self-tuned for each vertiex by analyzing the statistics of its neighborhood \cite{zelnik2004self}. From the adjacency matrix, we construct the diagonal degree matrix $\bm{D}$ and a graph Laplacian $\bm{L}$, using (\ref{eq:degree_mat}) and (\ref{eq:GL}), respectively. For the applications in this paper, we choose a normalized graph Laplacian with $p,\,q=0.5$.

\paragraph{Step 2 -- Selection Strategy.} 
Next, we describe the strategy for selecting $\Nh \ll \Nl$ nodes for which high-fidelity data is acquired. These nodes are chosen to be close to the centroids of the  clusters associated with the low-fidelity data $\bar{\mathcal{D}}$.  To find these, we compute the eigendecomposition of the graph Laplacian and leverage the properties of its low-lying eigenvalues and eigenfunctions. We embed each data point into the eigenfunctions space and apply a standard clustering algorithm (e.g. K-means) to determine the clusters and their centroids. The steps are,
\begin{enumerate}
    \item Compute the low-lying eigenfunctions of the graph Laplacian, $\bm{\phi}^{(m)},\,m=1,\,\dots,\,\Nh$.
    \item For every low-fidelity data point, $\lfdpi $, consider the coordinates in the eigenfunction space $\bm{\xi}^i \in \mathbb{R}^{\Nh}$. These are given by $ \bm{\xi}^i = [\phi_i^{(1)},\dots,\phi_i^{(\Nh)}], \; i = 1, \dots, \Nl$. 
    \item Use K-means clustering on the points $\{\bm{\xi}^i\}_{i = 1}^{\Nl}$ to find $\Nh$ clusters.
    \item For each cluster, determine the centroid and the low-fidelity data point closest to it.
    \item Re-index the low-fidelity snapshot and the parameter points so that the points identified above correspond to the first $\Nh$ points in the dataset.  
    \item Acquire high-fidelity data at the parameter values corresponding to these points, and assemble the data set $\mathcal{D}=\left\{\hfdpi \right\}_{i=1}^{\Nh}$, with $\hfdpi = \hfdp (\pari^i)$. Note that the elements of $\mathcal{D}$ are the high-fidelity counterparts of the first $\Nh$ elements of $\bar{\mathcal{D}}$.
    \item Scale the high-fidelity data with the same scaling factors used in Step 1 for the low-fidelity data. 
\end{enumerate}

\paragraph{Step 3 -- Bi-fidelity transformation.} In this step we generate a bi-fidelity approximation $\left\{ \mfdpi \right\}_{i=1}^{\Nl}$ that learns the data distribution from the low-fidelity dataset, and uses the select high-fidelity data to transform this distribution. 
The proposed bi-fidelity approach seeks a transformation that moves every low-fidelity data point to a new location in the data space, where the displacements are weighted sums of the $\Nh$ known displacements of the select points at which the high-fidelity counterpart is known.
That is,
\begin{eqnarray}
\mfdpi = \lfdpi + \sum_{j = 1}^{ \Nh } (\hfdpj - \lfdpj ) \psi^{(j)}_i, \qquad i=1,\,\dots,\,\Nl.
\label{eq:mftransf} 
\end{eqnarray}
Here $\mfdpi$ are the bi-fidelity data points, $\hfdpj - \lfdpj$ is the displacement vector that maps the $j$-th low-fidelity point to its high-fidelity location, and $\psi^{(j)}_i, j = 1, \dots, \Nh$, are the influence functions that determine the effect of the $j$-th displacement vector on the $i$-th point. We require the influence functions to encode the structure of the low-fidelity data distribution, and therefore a natural choice is to write them in terms of the eigenfunctions of the graph Laplacian. For consistency, we also require the influence functions to be a partition of unity. This is accomplished by applying a softmax activation to a set of auxiliary functions $v^{(j)}_i$ that are constructed as a linear combination of the low-lying eigenfunctions of the graph Laplacian. In particular, the influence functions are given by 
\begin{equation}
\psi^{(j)}_i = \frac{\exp{(v^{(j)}_i)}}{ \sum_{k = 1}^{\Nh} \exp{(v^{(k)}_i})}, \label{eq:defpsi} 
\end{equation}
and the auxiliary functions $v^{(j)}_i$ are,
\begin{equation}
    v^{(j)}_i = \sum_{m = 1}^{K} \alpha_{jm} \phi^{(m)}_i. \label{eq:defv} 
\end{equation}
The parameter $\alpha_{jm}$ determine the contribution of the $m$-th eigenfunction to the $j$-th auxiliary function, and $K$ denotes the cutoff in the spectrum of the graph Laplacian. This cutoff should be proportional to the number of high-fidelity data points. A suggested value, which is used in this study, is $K = 3 \Nh$.

The parameters $\bm{\alpha}=\{ \alpha_{jm} \}$ are determined by solving the minimization problem 
\begin{eqnarray}
\bm{\alpha}^* = \arg\min_{\bm{\alpha}} \, \mathsf{J} (\bm{\alpha}), \qquad \mathsf{J} (\bm{\alpha}) = \mathsf{J}_{\mathrm{data}}(\bm{\alpha}) + \omega \mathsf{J}_{\mathrm{reg}}(\bm{\alpha}).
\label{eq:defpi}
\end{eqnarray}

with

\begin{align}
     \mathsf{J}_{\mathrm{data}}(\bm{\alpha}) &= \frac{1}{\Nh} \sum_{i = 1}^{\Nh} || \mfdpi(\bm{\alpha}) - \hfdpi ||^2, \quad \mbox{ and} \label{eq:piterm-data}\\
     \mathsf{J}_{\mathrm{reg}}(\bm{\alpha}) &= \frac{1}{\tau^2 K \Nh}  \sum_{j = 1}^{\Nh}  \bm{v}^{(j)} (\bm{\alpha}) \cdot ( \bm{L} + \tau \bm{I} )^2 \bm{v}^{(j)} (\bm{\alpha}) .
     \label{eq:piterm-reg}
\end{align}

The first term in (\ref{eq:defpi}) is a data misfit term, which forces the bi-fidelity points to be close to the corresponding high-fidelity points. The second term is a structure-preserving term that promotes contributions from eigenfunctions with small eigenvalues. Its form is motivated by a similar term used in semi-supervised learning applications that utilize the graph Laplacian \cite{hoffmann2020consistency}. To examine the effect of this term, we substitute (\ref{eq:defv}) in (\ref{eq:piterm-reg}) to find 
\begin{equation}
\mathsf{J}_{\mathrm{reg}}(\bm{\alpha}) =  \frac{1}{K \Nh} \sum_{j=1}^{\Nh} \sum_{m=1}^{K}  \alpha_{jm}^2 \left(1 + \frac{\lambda_m}{\tau} \right)^2.  \label{eq:piterm-reg2}
\end{equation}
Thus, the values of $\alpha_{jm}$ corresponding to larger values of $\lambda_{m}$ are penalized more. Since the structure of the low-fidelity data is encoded in the eigenfunctions corresponding to the smaller eigenvalues, this term helps in carrying this structure over to the bi-fidelity model. It also makes the proposed algorithm relatively insensitive to the selection of the spectrum cutoff $K$, as the contribution from the higher-order eigenfunctions is weighed less. 

As discussed in Hoffmann et al. \cite{hoffmann2020consistency}, the presence of $\tau>0$ makes the minimization problem convex and easy to solve. A good candidate for its value is the smallest non-zero eigenvalue, i.e. $\tau=\lambda_2$. This amounts to solving a problem with a scaled spectrum of the Laplacian. 

It is also important to note that if we do not include the parameters $\pari$ in the definition of $\dpt$, i.e. $\dpt =  \qoi (\pari)$, they not appear explicitly in any of the equations. This means that the method can be applied to a generic point cloud $\{\lfdpi\}_{i=1}^{\Nl}$ embedded in $\mathbb{R}^D$ that has to be transformed based on a few more accurate, or updated, control points $\{\hfdpi\}_{i=1}^{\Nh}$, with $\Nh \ll \Nl$. This could represent a set of point-wise measurements that are dense in space but not very accurate, for which a smaller number of more precise measurements are available.

The regularization constant $\omega$ in (\ref{eq:defpi}) balances the interplay between the data misfit term and the structure-preserving regularization term. If its value is too small, the bi-fidelity model is likely to over-fit the high-fidelity data and ignore the structure learned from the low-fidelity data. 
On the other hand, if its value is too large, the method will yield bi-fidelity points that are significantly further way from their high-fidelity counterpart. As described in Appendix \ref{appx_Lcurve}, the optimal value of this parameter may be determined using the L-curve method \cite{hansen1993use}.

\section{Theoretical Analysis}
\label{sec:theory} 
In this section we summarize some desirable theoretical properties of the proposed method. These results are derived in Appendix \ref{appx_gradient}, \ref{appx_hessian}, and \ref{appx_canonical}. 

\paragraph{Property 1} An explicit expression for the gradient of the loss function $\mathsf{J}(\bm{\alpha})$ with respect to the optimization parameters $\bm{\alpha}$ can be computed, lowering the computational cost of the algorithm.

\paragraph{Property 2} In the limit of a small data misfit term (which happens as the optimization iterations converge), the Hessian of the loss function is positive definite. This proof is based on recognizing that (\emph{i}) the data misfit term is in the form of a least-squares residual and (\emph{ii}) the regularization term is a positive-definite quadratic form. This ensures that the resulting optimization problem is solved easily.

\paragraph{Property 3} Under the assumptions (\emph{a}) the low-fidelity data is partitioned into $M$ clusters, (\emph{b}) the high-fidelity data differs from the low-fidelity data by distinct rigid translations applied to each cluster, and (\emph{c}) the high-fidelity version of one point per cluster is known, the proposed approach permits a transformation that maps each low-fidelity point to the true high-fidelity point in the limit of infinite low-fidelity data and as the regularization parameter tends to zero. That is, the bi-fidelity data set converges to its high-fidelity counterpart. This is a consistency results that demonstrates the proposed method can solve this special problem exactly. This is described in more detail in Appendix \ref{appx_canonical}.

\section{Numerical Experiments} \label{num-experiment}
In this section we apply the proposed method to a collection of numerical problems. The first problem is an illustrative example in a two-dimensional data space, where each step can be easily visualized. The second problem is an application to linear elasticity, where the data space dimension is higher, with $D=5$, while the third is a fluid dynamics problem, with $D = 3$. 

\subsection{Illustrative example}
\begin{figure}
\centering
\begin{subfigure}[b]{0.25\textwidth}
\centering
\includegraphics[height=.95\textwidth]{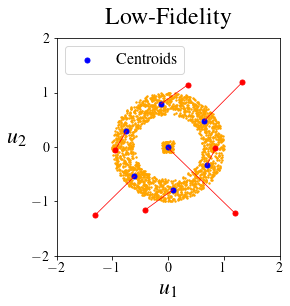}
\caption{}
\label{fig:bullsEye_a}
\end{subfigure}%
\begin{subfigure}[b]{0.25\textwidth}
\centering
\includegraphics[height=.95\textwidth]{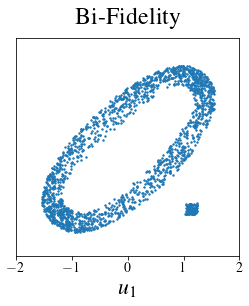}
\caption{}
\label{fig:bullsEye_b}
\end{subfigure}%
\begin{subfigure}[b]{0.25\textwidth}
\centering
\includegraphics[height=.95\textwidth]{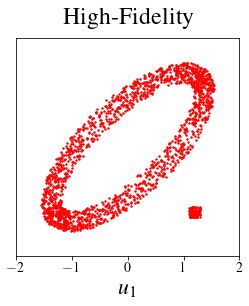}
\caption{}
\label{fig:bullsEye_c}
\end{subfigure}
\begin{subfigure}[b]{\textwidth}
\centering
\includegraphics[width=.69\textwidth]{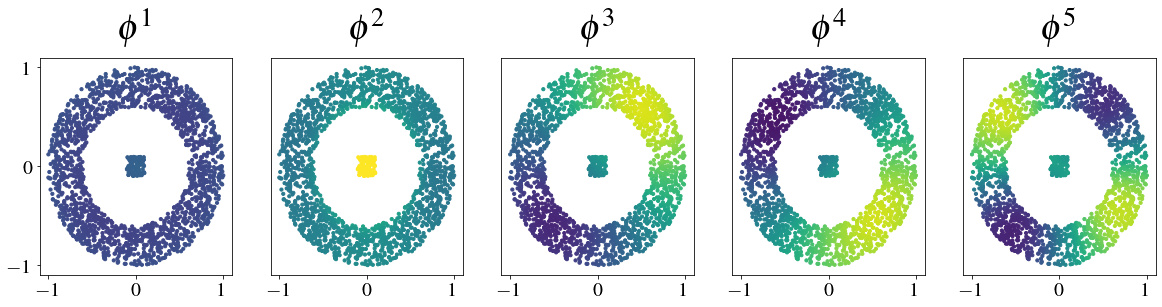}
\caption{}
\label{fig:bullsEye_d}
\end{subfigure}
\begin{subfigure}[b]{\textwidth}
\centering
\includegraphics[width=.69\textwidth]{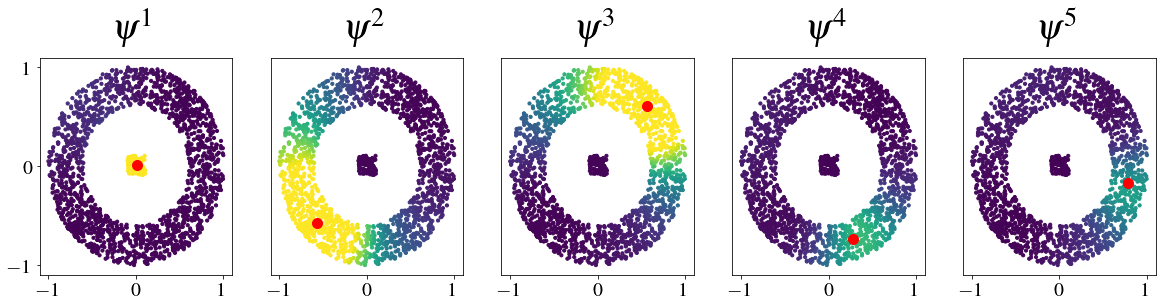}
\caption{}
\label{fig:bullsEye_e}
\end{subfigure}
\caption{(a) Low-fidelity data distribution. The blue points are the closest to the centroids, and the red points are their high-fidelity counterparts. (b) Bi-fidelity correction of the low-fidelity data. (c) High-fidelity data distribution. (d) The first five eigenfunctions corresponding to the low-lying spectrum of the graph Laplacian. (e) The first five influence functions. }
\end{figure}

We consider the low- and high-fidelity datasets shown in Figures \ref{fig:bullsEye_a} and \ref{fig:bullsEye_c}. 
A visual inspection of the low-fidelity data reveals a structure comprising a circle surrounded by a ring. In the high-fidelity data this structure is modified so that circle translates to a position outside the ring, while the ring is stretched via an affine transform. 
The goal is to construct a bi-fidelity approximation using the low-fidelity dataset (with $\Nl = 2,000$ points) and a few select high-fidelity points ($\Nh = 7$). 

First, the graph Laplacian $\bm{L}$ is constructed using the low-fidelity data with $\sigma = 0.25$ and $p,\,q = 0.5$. Thereafter, the low-fidelity data points are embedded in the eigenfunctions space, and a K-means algorithm is used to find $\Nh=7$ clusters and their centroids, and the low-fidelity data points closest to these centroids. 
In Figure \ref{fig:bullsEye_a} these points, indicated by blue dots, are paired with their high-fidelity counterparts, indicated by red dots, and connected via a line segment. 
These pairs are used to define the data misfit loss term in Equation (\ref{eq:piterm-data}).

The low-lying eigenfunctions are plotted in Figure \ref{fig:bullsEye_d}.
As expected, the first eigenfunction is constant, the second (the Fielder vector) isolates the circle from the ring, while the other three split the ring into structures with smaller scale. 

In the final step, the minimization problem (\ref{eq:defpi}) is solved to determine the influence functions and the bi-fidelity dataset. This is done with $\tau=\lambda_2=0.005$, $\omega=10^{-6}$, and a spectrum cutoff of $K=20$. The first five influence functions are displayed in Figure \ref{fig:bullsEye_e}. We observe that the influence function corresponding to the data point in the inner circle attains a value close to unity on the circle and zero on the ring. This implies that in the bi-fidelity update, all points in the inner circle will move together with the point for which the high-fidelity data is available. The second and third influence functions attain a large value at the points within the ring where the high-fidelity data is known. Their influence decays as one moves away from these points. This implies that they will cause their respective halves of the ring to move along with the point for which the high-fidelity data is available. Since the motion of the high-fidelity data points is outward, this will lead to a stretching of the outer circle. 
The comparison in Figures \ref{fig:bullsEye_b}-\ref{fig:bullsEye_c} shows that the final bi-fidelity dataset closely matches the structure of the high-fidelity dataset. 

\subsection{Traction on a soft material with a stiff inclusion}
\label{sec:tracion_prob}
\begin{figure}
\centering
\begin{subfigure}[b]{.4\textwidth}
  \centering
  \includegraphics[width=.75\textwidth]{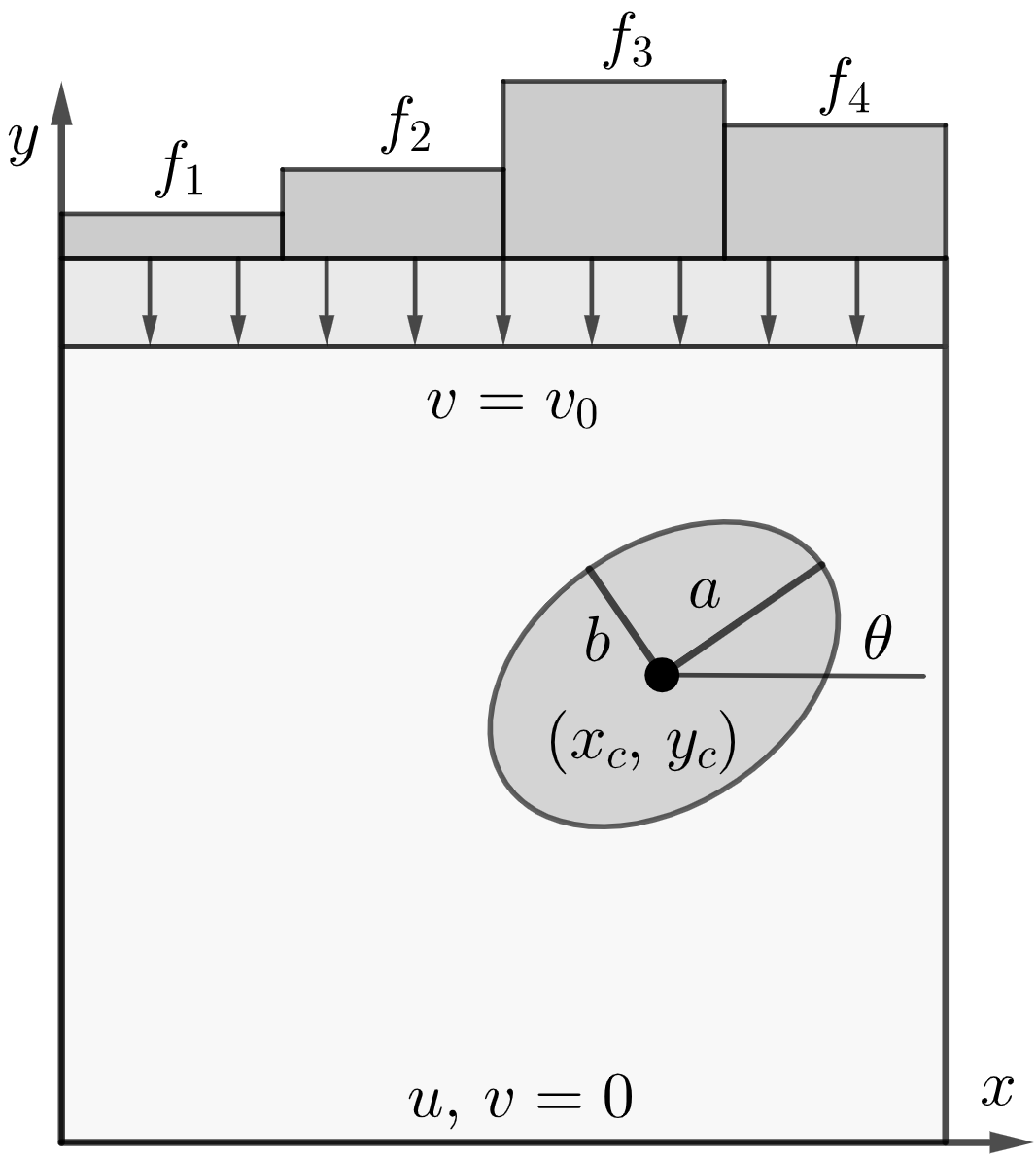}
  \caption{}
  \label{fig:inclusion_schematic}
\end{subfigure}%
\hfill
\begin{subfigure}[b]{.6\textwidth}
  \centering
  \includegraphics[width=.95\textwidth]{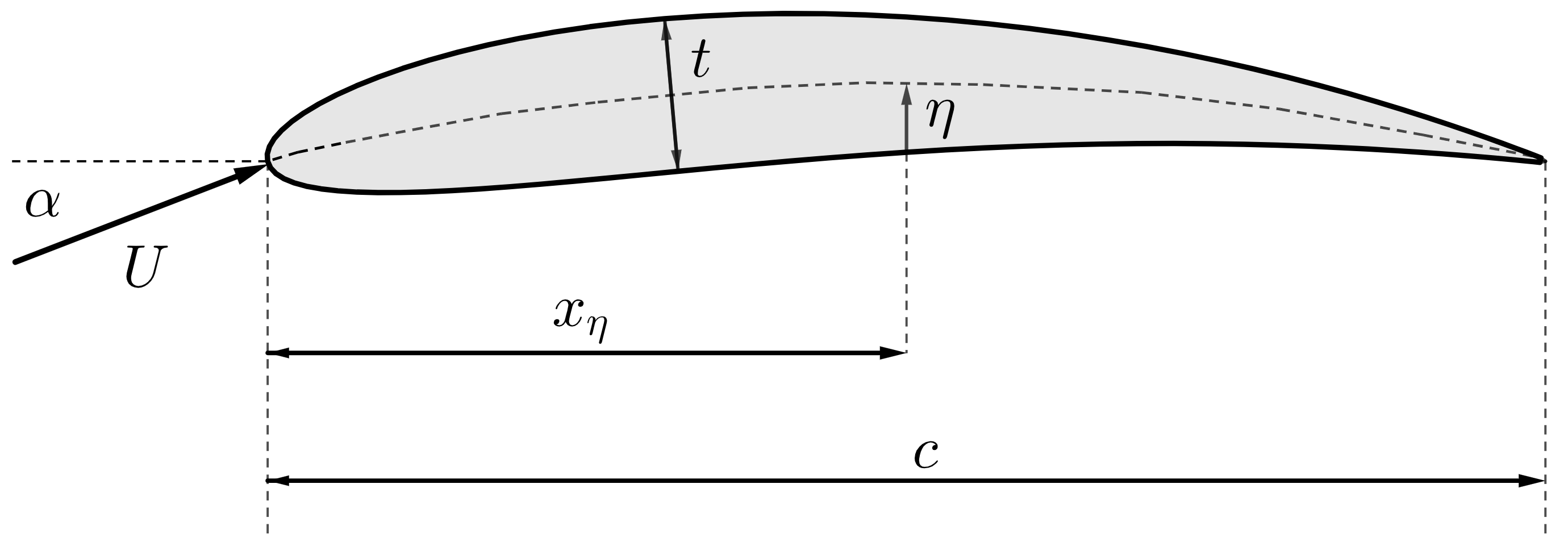}
  \caption{}
  \label{fig:airfoil_schematic}
\end{subfigure}
\caption{(a) Schematic of the soft body (light grey) with the elliptic stiffer inclusion (dark grey). The square is compressed on top with a uniform displacement $v=v_0$, while the bottom is fixed. The vertical traction is integrated over the top side across equal sections to compute the localized forces $f_i$. (b) Schematic of the airfoil with the input parameters of the problems.}
\end{figure}

\paragraph{Problem description.} 
We examine a problem of linear elasticity which involves a soft square sheet in plane stress with an internal stiffer elliptic inclusion. The length of the edge of the square is $L =10 \, \mathrm{cm}$, its Young's modulus $E = 1 kPa$, and the Young's modulus of the inclusion is $E_i = 4E$. Both the body and the inclusion are incompressible. The lower edge of the square is fixed, and a uniform downward displacement $v_0 = -5 \, \mathrm{mm}$ is applied to the upper edge. The top edge is traction free in the horizontal direction while the vertical edges are traction-free in both directions (Fig. \ref{fig:inclusion_schematic}). We wish to predict attributes of the vertical traction field on the upper edge as a function of the inclusion shape, orientation and location. This problem is motivated by the desire to identify stiff tumors within a soft background tissue, which is especially relevant to detecting and diagnosing breast cancer tumors \cite{sarvazyan2012mechanical,barbone2010review}.  
 
\paragraph{Parameters and Quantities of Interest.} 
The input parameters of the problem are the coordinates of the center of the elliptical inclusion $(x_c,\,y_c)$, its orientation $\theta$, and its major and minor semi-axes $a$ and $b$ (see Fig. \ref{fig:inclusion_schematic}). The minimum and maximum values for these parameters are reported in Table \ref{tab:Params_range}. The output quantities of interest include the values of the localized vertical forces on the upper edge. These are determined by dividing the top edge into $M = 4$ sections of equal length and integrating the vertical traction $\sigma_{yy}(x,\,y)$ over each section. This results in $M$ values of localized forces $f_i, \; i = 1, \dots, M$ (see Fig. \ref{fig:inclusion_schematic}),
\begin{equation}
    f_i = \int_{(i-1)\frac{L}{M}}^{i\frac{L}{M}} \sigma_{yy} (x,\, L) \mathrm{d}x,\;\;\;\;i=1,\,\dots,\,M. 
\end{equation}

In addition to these forces, we include the maximum value of traction on the top edge as another quantity of interest. 
Therefore, the $M+1$ quantities of interest are $q_i = f_i, \; i=1,\,\dots,\,M$, and $q_{M+1} = \max_x \sigma_{yy}(x,\,L)$. As the location, orientation and size of the inclusion is varied, the traction field on the top surface changes, which in turn changes the $M$ components of the localized force, and the maximum value of traction.

We consider the case where the data space is constructed only from the output vector, that is $\dpt(\pari) = \qoi(\pari)$. The case of including the input vector in the data space, that is, $\dpt(\pari) = [\pari, \qoi (\pari)]$ is described in Appendix \ref{appx_dataspace}.

\begin{table}[hbt!]
\centering
\begin{tabular}{|c c c c|}
\hline 
\multicolumn{4}{|c|}{\textbf{Soft body with inclusion}}\tabularnewline
\hline
Parameter & Minimum & Maximum & Units\\
\hline
\rule{0pt}{2ex} $x_c$ & 2.5 & 7.5 & cm\\
\rule{0pt}{2ex} $y_c$ & 5 & 7.5 & cm\\
\rule{0pt}{2ex} $\theta$ & 0 & 180 & degree\\
\rule{0pt}{2ex} $a$ & 1 & 2 & cm\\
\rule{0pt}{2ex} $b$ & 1 & 2 & cm\\
\hline
\end{tabular}%
\hspace{1cm}
\begin{tabular}{|c c c c|}
\hline 
\multicolumn{4}{|c|}{\textbf{Airfoil}}\tabularnewline
\hline
Parameter & Minimum & Maximum & Units\\
\hline
\rule{0pt}{2ex} $\eta$ & 1 & 6 & \%c\\
\rule{0pt}{2ex} $x_\eta$ & 4 & 7 & $0.1 \times c$\\
\rule{0pt}{2ex} $t$ & 10 & 20 & \%c\\
\rule{0pt}{2ex} $\alpha$ & -5 & 12 & degree\\
\rule{0pt}{2ex} $Re$ & $10^3$ & $10^7$ & 1\\
\hline
\end{tabular}
\vspace{2mm}
\caption{\label{tab:Params_range} Ranges spanned by input parameters for the traction and airfoil problems.}
\end{table}

\paragraph{Low- and High-Fidelity models.} 
We employ two finite element-based models that differ in the number of elements of the mesh. The low-fidelity model uses a coarse mesh with around $400$ finite elements, whereas the high-fidelity model has a fine mesh with around $25,000$ elements. It is verified that the high-fidelity model produces a solution that is mesh-converged. 

\paragraph{Results.} We generate $\bar{N} = 4084$ samples of the input parameters by treating each parameter as an independent random variable that is uniformly distributed within its range. 
For each instance of parameters we generate the low- and high-fidelity solutions. We use $\Nh=30$ high-fidelity data points to construct the bi-fidelity results, and the remainder for testing the performance of method. We observe that the low-fidelity solutions can capture the trend of the traction field, but generally underestimate its magnitude (see Appendix \ref{appx_traction}).

\begin{figure}
\centering
\begin{minipage}{.56\textwidth}
\hfill
\begin{subfigure}[b]{\textwidth}
\hfill
  \centering
  \includegraphics[width=0.32\textwidth]{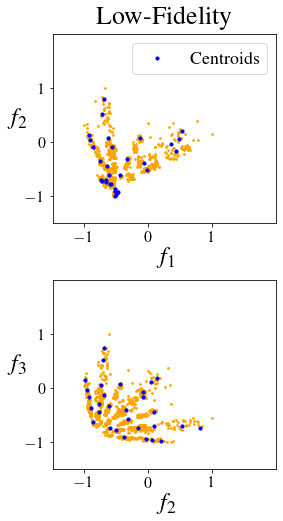}
  \includegraphics[width=0.32\textwidth]{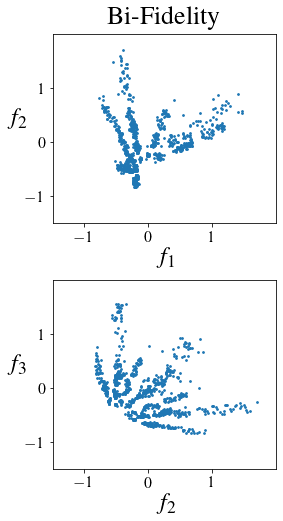}
  \includegraphics[width=0.32\textwidth]{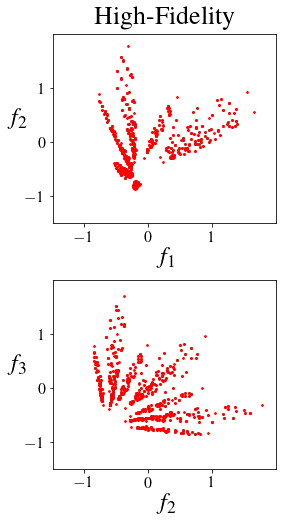}
  \caption{}
  \label{fig:traction_results_a}
\end{subfigure}
\begin{subfigure}[b]{\textwidth}
  \centering
  \includegraphics[width=0.95\textwidth]{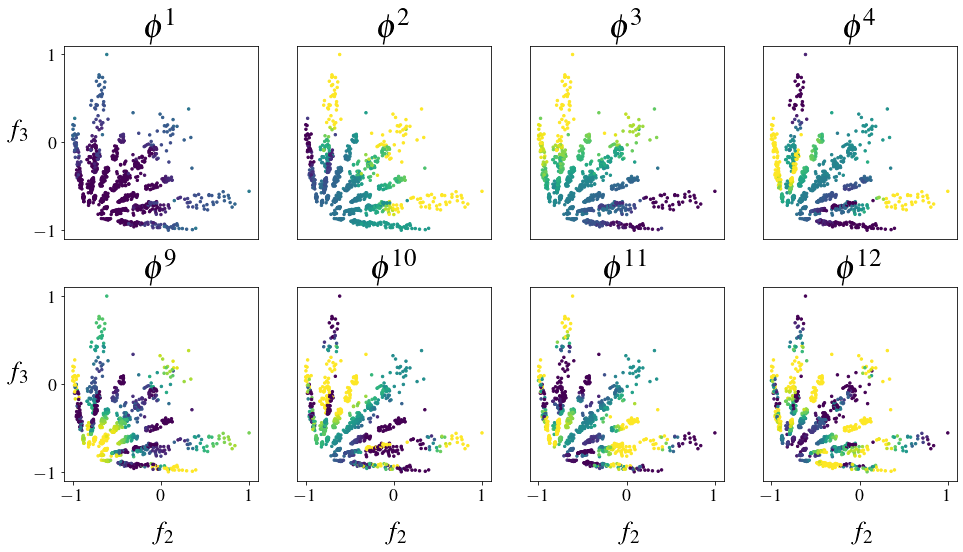}
  \caption{}
\label{fig:traction_results_b}
\end{subfigure}
\begin{subfigure}[b]{\textwidth}
  \centering
  \includegraphics[width=0.95\textwidth]{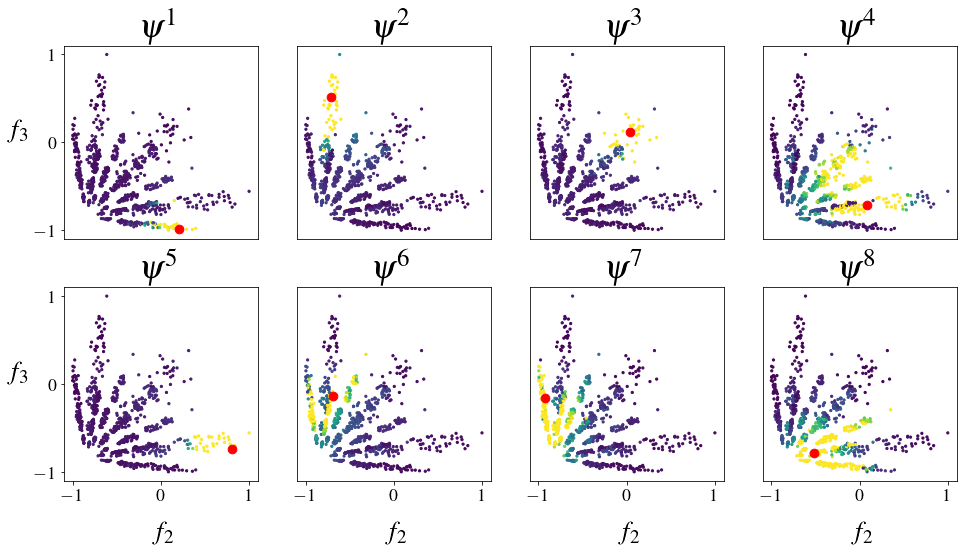}
  \caption{}
  \label{fig:traction_results_c}
\end{subfigure}
\end{minipage}%
\begin{minipage}{.425\textwidth}
\begin{subfigure}[b]{\textwidth}
\vspace{0.4cm}
\centering
\hfill
  \includegraphics[width=0.925\textwidth]{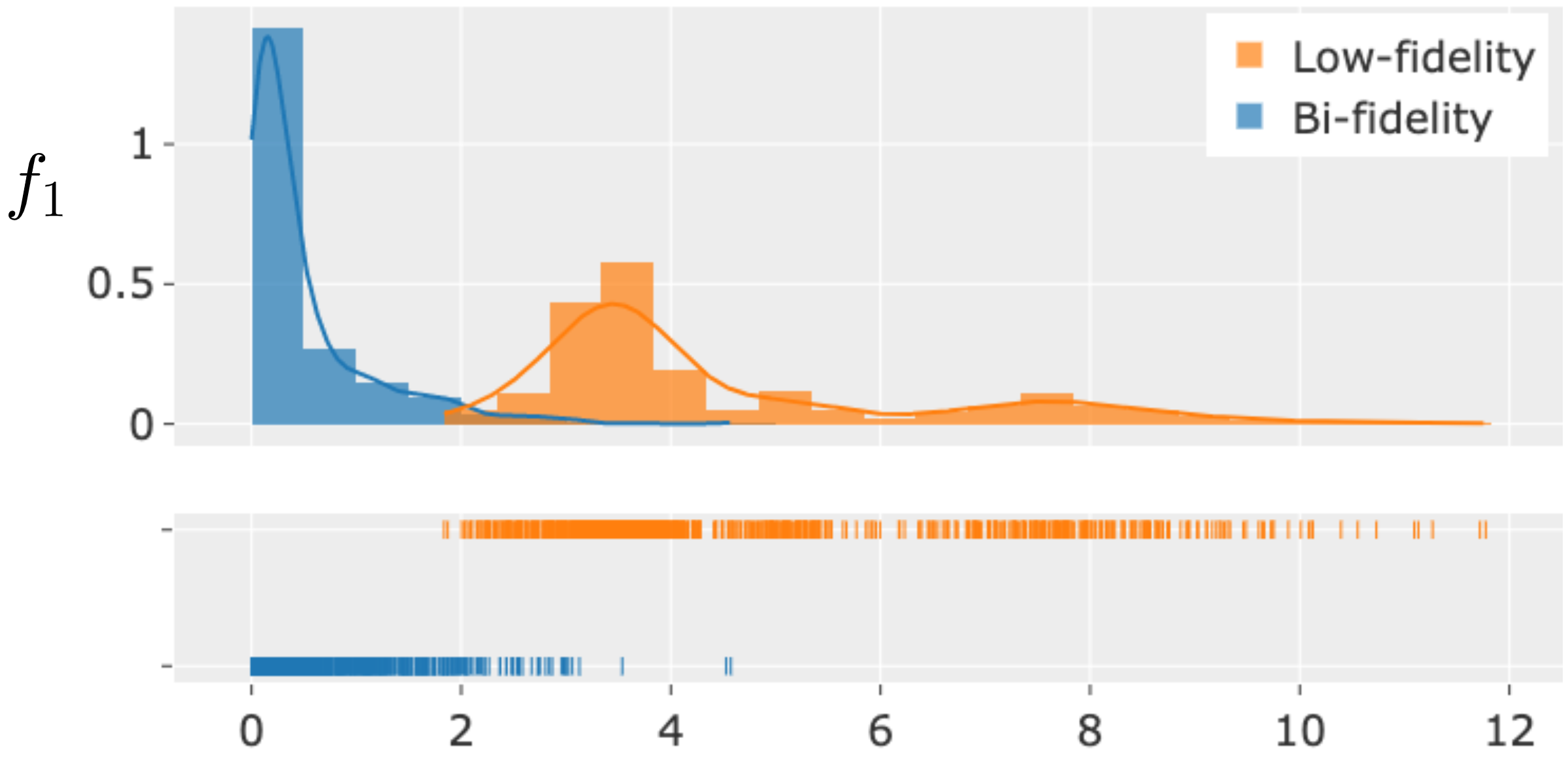}
  \vspace{0.1cm}
  \hfill
  \includegraphics[width=0.925\textwidth]{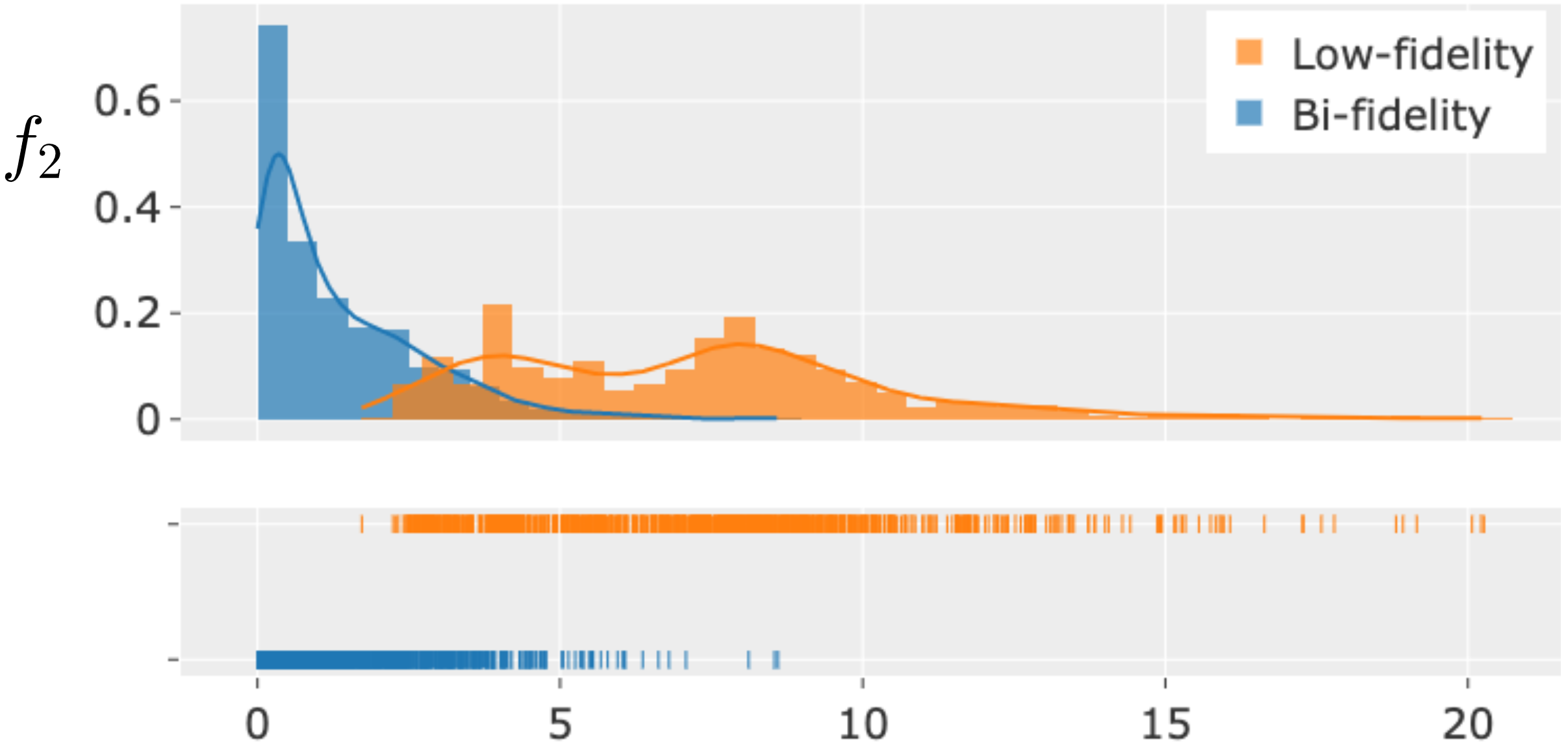}
  \vspace{0.1cm}
  \hfill
  \includegraphics[width=0.925\textwidth]{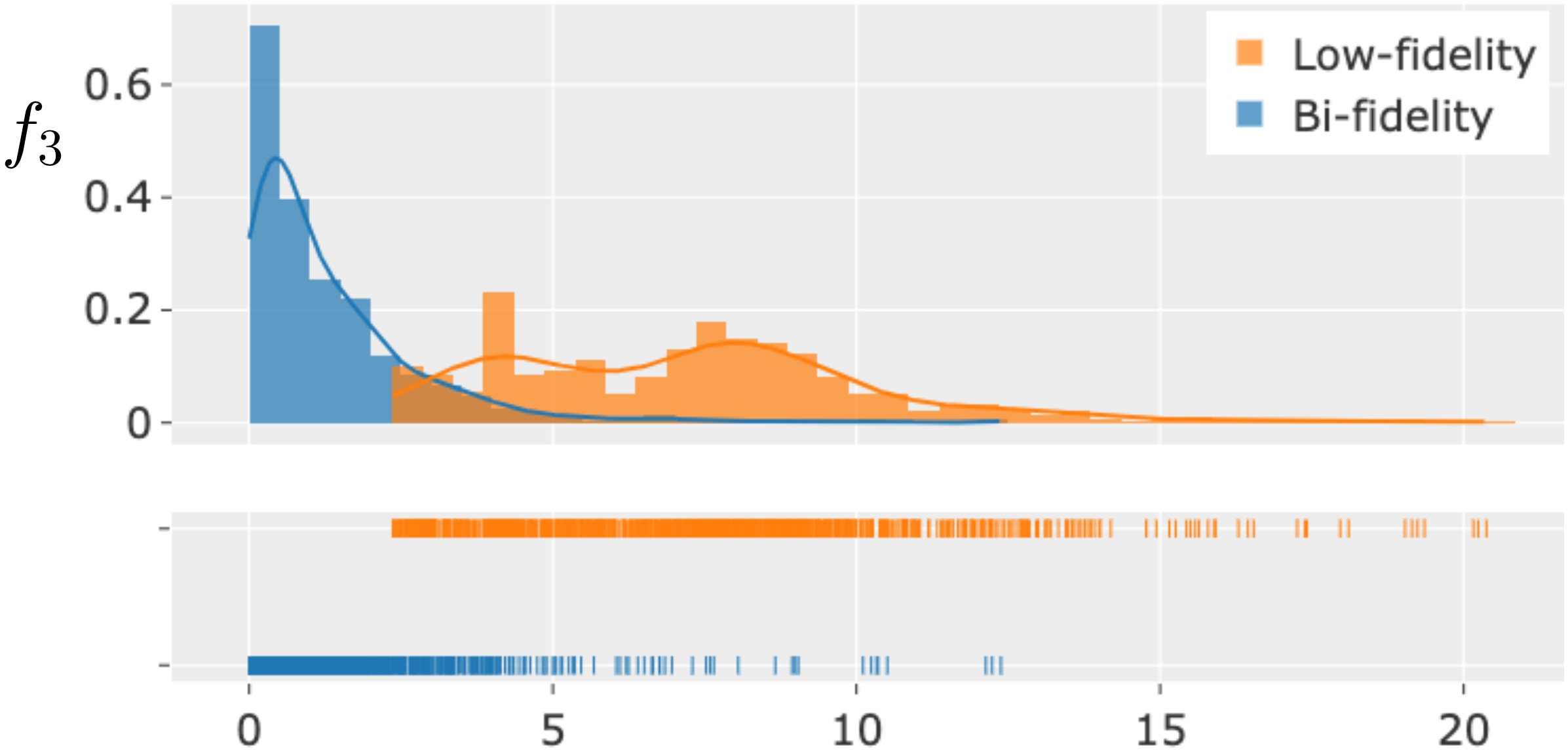}
  \vspace{0.1cm}
  \hfill
  \includegraphics[width=0.925\textwidth]{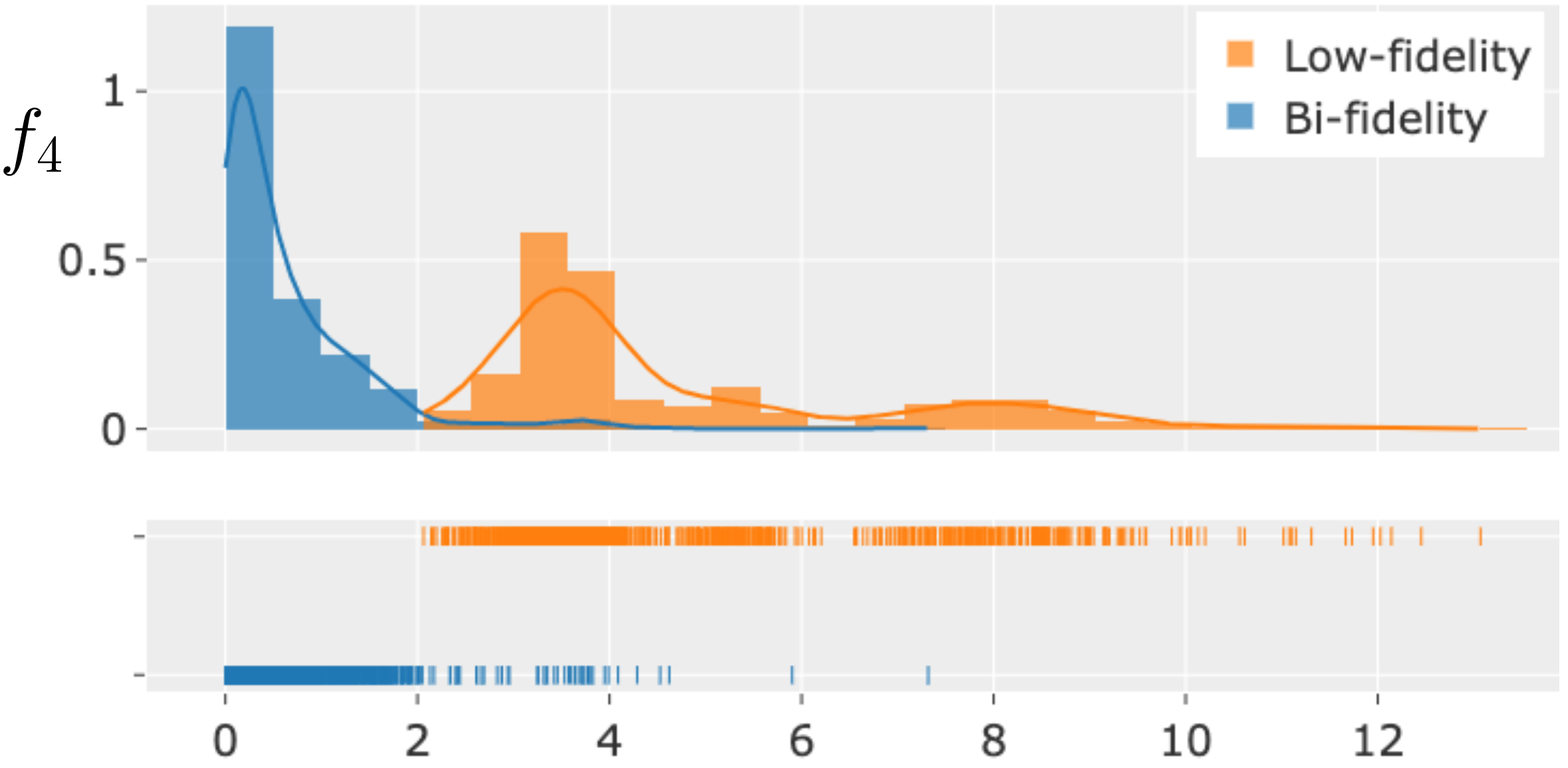}
  \vspace{0.1cm}
  \hfill
  \includegraphics[width=0.975\textwidth]{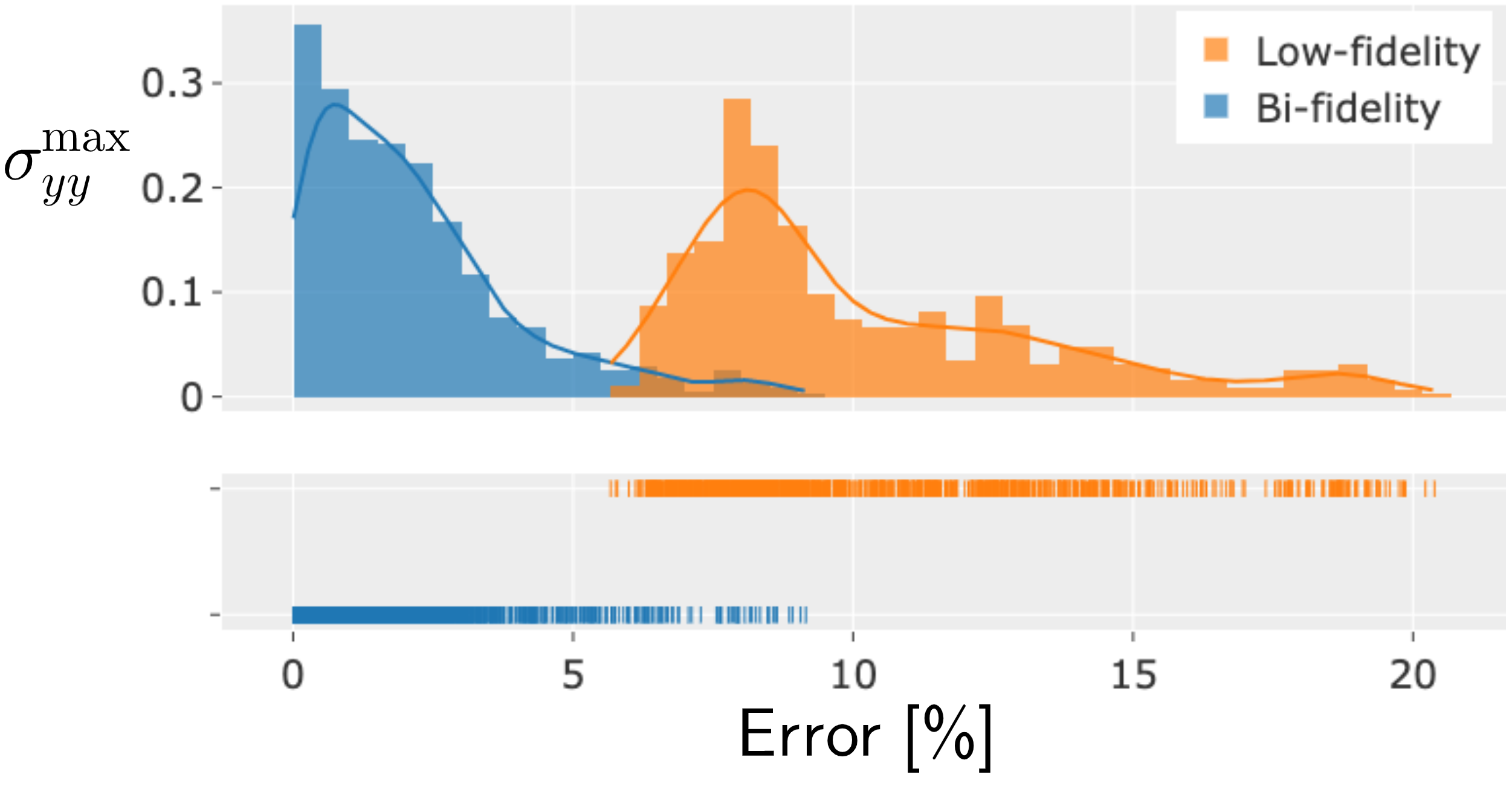}
  \caption{}
  \label{fig:traction_results_d}
\end{subfigure}
\end{minipage}
\caption{Results for the traction problem. (a) Comparison of the low-, bi- and high-fidelity data. The low-fidelity data are shown together with the points closest to the centroids of the clusters (in blue). (b) Eight eigenfunctions from the low-lying spectrum  projected onto the $(f_2,\,f_3)$ plane. (c) Influence functions for eight control points projected onto the $(f_2,\,f_3)$ plane. (d) Error distribution for the low- and bi-fidelity model for each output component.}
\end{figure}

To visualize the five-dimensional data set, we project the data points in the $(f_1,\, f_2)$ and $(f_2,\, f_3)$ planes. 
In Figure \ref{fig:traction_results_a}, we show a comparison between the scaled low- and high-fidelity datasets. We notice that similar structures arise in both datasets, however the magnitude of the low fidelity data is generally smaller.

The graph Laplacian is constructed from the low-fidelity data points with $p,q=0.5$ and a self-tuned $\sigma$.
The low-lying eigenfunctions are shown in Fig. \ref{fig:traction_results_b} in the $(f_2,\,f_3)$ plane. We observe that the eigenfunctions corresponding to larger eigenvalues tend to have higher-frequency fluctuations and finer spatial structure. Following Step 2, we determine the coordinates of each low-fidelity data point in the eigenfunctions space and then perform K-means clustering to find the points closest to the centroids of $\Nh=30$  clusters. These points are shown in leftmost plots of Figure \ref{fig:traction_results_a}, and correspond to the only points where we assume that the high-fidelity data is available. We observe that these points appear to be evenly distributed over the span of the low-fidelity data.

Next, following Step 3, we determine the coefficients for the influence functions in the bi-fidelity approximation. In solving the minimization problem (\ref{eq:defpi}), we choose a value of $\tau=5\cdot 10^{-10}$, which corresponds to the smallest non-zero eigenvalue, and a spectrum cutoff of $K=90$. To select the value of the regularization parameter we make use of the L-curve method, and study the data misfit loss versus the regularization loss for different values of $\omega\in[10^{-7},\,10^{-3}]$. The optimal value corresponding to the elbow of the curve is found to be $\omega^*=3.8 \cdot 10^{-6}$. The solution of the minimization problem leads to an influence function for each point where the high-fidelity version is known. A sample of these functions and the corresponding data point is shown in Fig. \ref{fig:traction_results_c}, in the $(f_2,\,f_3)$ plane. We observe that all influence functions peak at their respective data point and vanish away from it.

The final bi-fidelity data approximation, which is generated via (\ref{eq:mftransf}), is shown in the middle plot of Figure \ref{fig:traction_results_a}. When comparing the low-, high-, and bi-fidelity data points we observe that bi-fidelity data distribution is closer to the high-fidelity distribution. The predominant effect of the bi-fidelity transformation is to apply a stretch to the low-fidelity data so that it better matches the high-fidelity data at the points where these are available. 

To quantify the error  in the low- and bi-fidelity data, we compute the relative absolute difference with respect to the high-fidelity data at each point $i$ and component $k$,
\begin{equation}
    \bar{e}_k^i =  \frac{| \bar{u}^i_k - u^i_k |}{ E(|u^i_k|)} \times 100\%, \quad 
    e_k^i = \frac{| w^i_k - u^i_k |}{ E(|u^i_k|)} \times 100\%, 
    \qquad i=1,\,\dots,\,N_{val}, \quad k=1,\,\dots,\,D.
    \label{eq:error}
\end{equation}
where $E(\cdot)$ denotes the average over all points $i$, and $N_{val}$ is the number of validation points (in this case, $N_{val} = \Nl - \Nh$).
The mean value of these errors, i.e. $E(\bar{e}_k^i)$ and $E(e_k^i)$, are reported in Table \ref{tab:errors} for each component, and their distribution is shown in Figure \ref{fig:traction_results_d}. 
We observe that the error of the low-fidelity data ranges between 5-10\%, while that of the bi-fidelity data is around or below 2\%, indicating an improvement of 5-9 times. The histograms and the distribution plots for the error shown in Figure \ref{fig:traction_results_d} provide more detail. In each case we observe that the error distribution for the bi-fidelity data is more closely centered around zero and presents a much smaller spread when compared to the distribution of the low-fidelity errors. 

\begin{table}[hbt!]
\centering
\begin{tabular}{|c|c c c c c|c c c|}
\cline{1-9} 
\multicolumn{1}{|c|} {\textbf{Error [\%]}} & \multicolumn{5}{c|}{\textbf{Soft body with inclusion}} & \multicolumn{3}{c|}{\textbf{Airfoil}}\tabularnewline
\hline 
\rule{0pt}{3ex} Quantity of interest & $f_{1}$ & $f_{2}$ & $f_{3}$ & $f_{4}$ & $\sigma_{yy}^{\mathrm{max}}$ & $C_{L}$ & $C_{D}$ & $C_{M}$\tabularnewline
\hline 
\rule{0pt}{3ex} Low-fidelity & 4.48 & 7.15 & 7.21 & 4.65 & 10.19 & 42.79 & 28.81 & 216.11 \tabularnewline
\rule{0pt}{3ex} Bi-fidelity & 0.50 & 1.35 & 1.37 & 0.63 & 2.13 & 19.90 & 11.71 & 46.38 \tabularnewline
\rule{0pt}{3ex} Improvement factor & 8.96 & 5.22 & 5.19 & 6.94 & 4.63 & 2.15 & 2.46 & 4.66 \tabularnewline
\hline 
\end{tabular}
\vspace{2mm}
\caption{\label{tab:errors} Error of the low- and bi-fidelity model for each quantity of interest.}
\end{table}

\subsection{Aerodynamic coefficients of NACA airfoils}
\label{sec:airfoil}
\paragraph{Problem description.} 
The bi-fidelity approach is now used to tackle a problem of aerodynamics where the goal is to predict the lift, drag and pitching moment coefficients for a family of airfoils operating at different regimes and conditions. We consider the 4-digit NACA airfoils, whose shape is defined by 3 geometric parameters, and investigate how the aerodynamic performance of these airfoils changes at different Reynolds numbers and angles of attack. 

\paragraph{Parameters and Quantities of Interest.} 
The parameters of the problem comprise both design and operating condition variables. These are the maximum camber of the airfoil $\eta$, the distance of the maximum camber from the airfoil leading edge $x_\eta$, the thickness of the airfoil $t$, the angle of attack $\alpha$, and the Reynolds number $Re = \frac{U c}{\nu}$ (see Figure \ref{fig:airfoil_schematic}). Here, $U$ is the flow speed, $c$ is the chord of the airfoil (unitary in our case), and $\nu$ is the kinematic viscosity of the fluid. In our analysis, the Reynolds number is varied by changing the flow speed $U$.
The range of each parameter is reported in Table \ref{tab:Params_range}.

The quantities of interest are the aerodynamic coefficients $C_L,\,C_D$, and $C_M$, defined as,
\begin{eqnarray}
C_L = \frac{L}{\frac{1}{2}\rho U^2 c}, \qquad C_D = \frac{D}{\frac{1}{2}\rho U^2 c}, \qquad C_M = \frac{M}{\frac{1}{2}\rho U^2 c^2},
\label{eq:aeroCoeffs}
\end{eqnarray}
where $L,\,D$ and $M$ are the lift, drag and the pitching moment about a point located at quarter chord from the leading edge, respectively, and $\rho$ is the density of the fluid. Hence, the vector of quantities of interest is $\qoi(\pari) = [C_L, \, C_D, \, C_M]$, with input parameters $\pari=[\eta,\, x_\eta,\, t,\, \alpha,\, Re]$. The data space is formed by the output quantities of interest only, i.e. $\dpt(\pari) = \qoi(\pari)$. A different case, where the Reynolds number is included in the data space, is analyzed in Appendix \ref{appx_dataspace}.

\paragraph{Low- and high-fidelity model.}
The low-fidelity data is generated using XFOIL \cite{xfoil89}, a code based on the vortex panel method for the analysis of subsonic airfoils. In this case, the lift and moment coefficients are calculated by direct integration  of surface pressure, whereas the drag is recovered by applying the Squire-Young formula \cite{SquireYoung}.
To generate the low-fidelity results, the surface of each airfoil is discretized with 40 panels, and the Reynolds and Mach numbers are set by using
a kinematic viscosity of $\nu =10^{-5} \, \mathrm{m^2\,s^{-1}}$ and speed of sound $c_{\mathrm{s}} = 340 \; \mathrm{m \; s^{-1}}$.

The high-fidelity results are generated via 2D Reynolds-averaged Navier–Stokes (RANS) simulations with a $\mathrm{SST} \;\; k-\omega$ turbulence model \cite{menter1993zonal} using OpenFOAM.
The computational domain is a cuboid of dimension $1000 c \times 1000 c \times c$. We use a hybrid mesh, comprising a C-grid structured mesh in the proximity of the airfoil of size $4c\times6c$, and an unstructured mesh in the rest of the domain (see Appendix \ref{appx_airfoil}). 
The number of finite volumes of the mesh varies between 100,000 and 800,000, depending on the Reynolds number.
At the outer boundary we apply Dirichlet boundary conditions for both velocity and pressure, while
on the airfoil surface we use no-slip condition for velocity and zero-gradient condition for pressure.
The turbulence intensity of the flow at the outer boundary is set to 2\%. 

\begin{figure}
\centering
\begin{subfigure}[b]{.6\textwidth}
  \centering
  \includegraphics[width=0.95\textwidth]{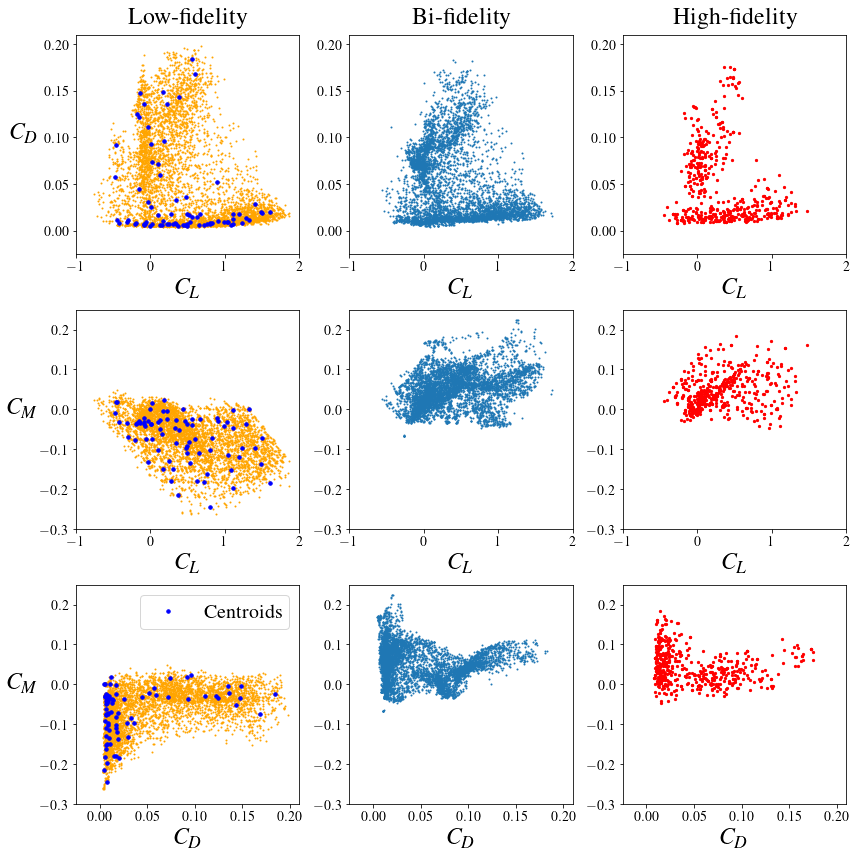}
  \vspace{0.25cm}
  \caption{}
  \label{fig:airfoil_results_a}
\end{subfigure}%
\begin{subfigure}[b]{.39\textwidth}
  \centering
  \includegraphics[width=0.95\textwidth]{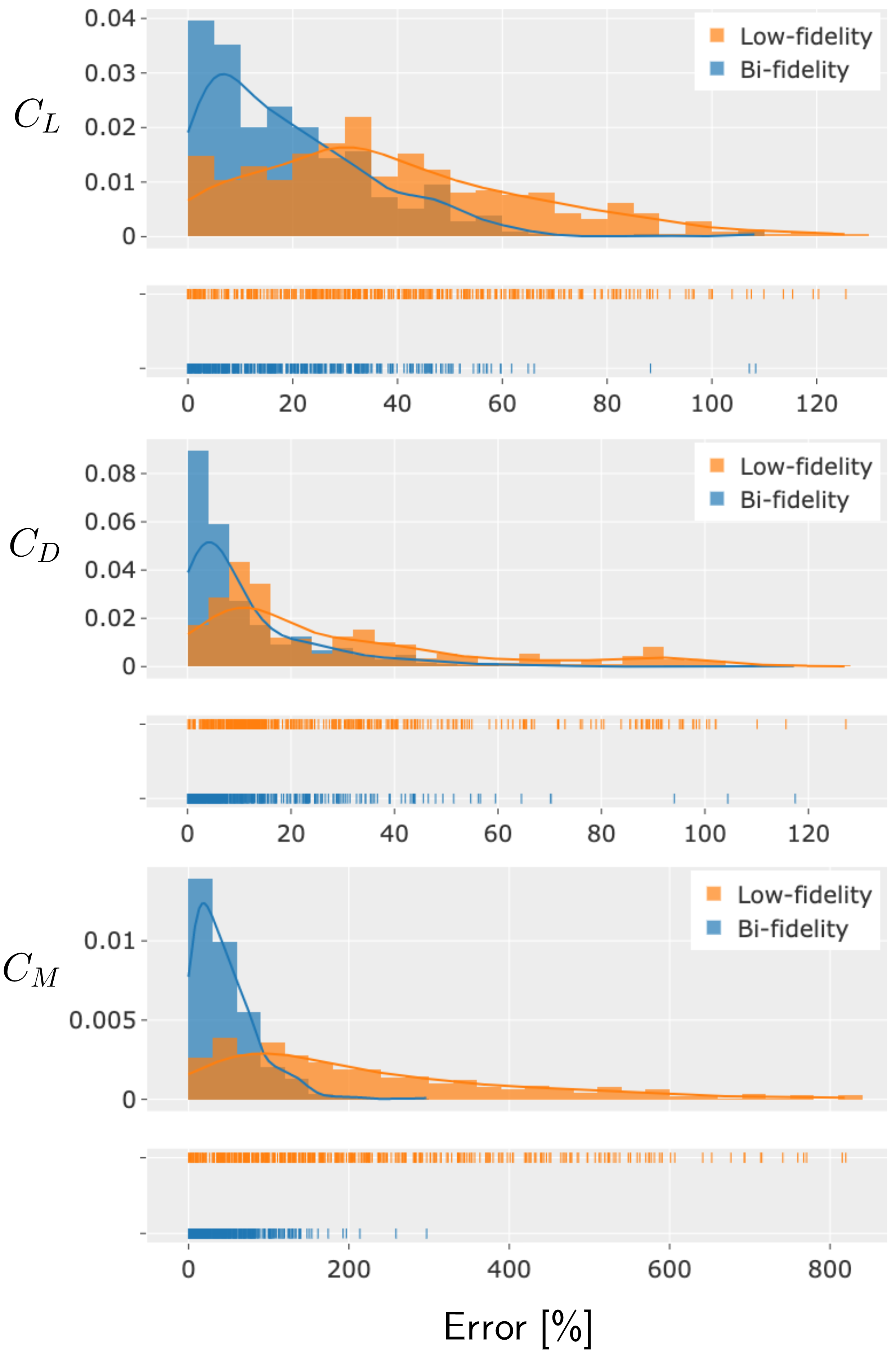}
  \caption{}
  \label{fig:airfoil_results_b}
\end{subfigure}
\begin{subfigure}[b]{0.8\textwidth}
\vspace{0.5cm}
  \centering
  \includegraphics[width=0.32\textwidth]{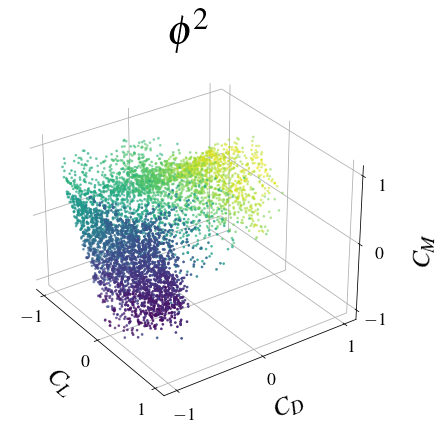}
  \includegraphics[width=0.32\textwidth]{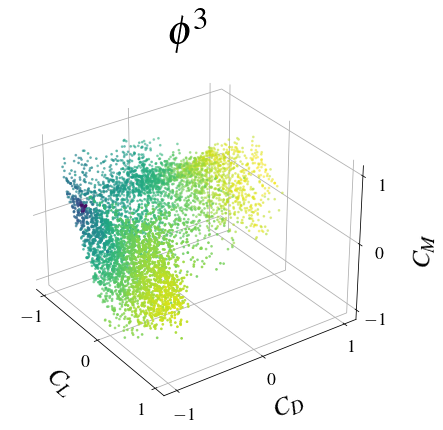}
  \includegraphics[width=0.32\textwidth]{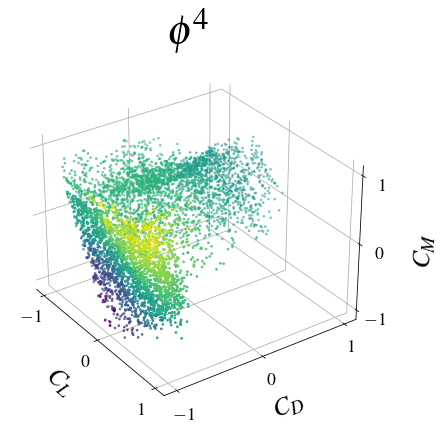}
  \caption{}
  \label{fig:airfoil_results_c}
\end{subfigure}
\begin{subfigure}[b]{0.8\textwidth}
  \includegraphics[width=0.32\textwidth]{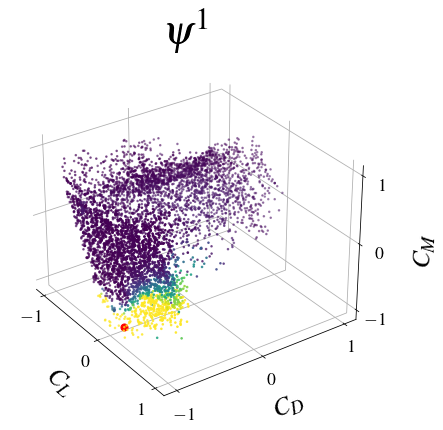}
  \includegraphics[width=0.32\textwidth]{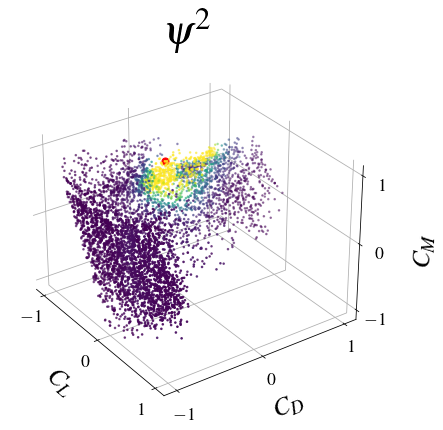}
  \includegraphics[width=0.32\textwidth]{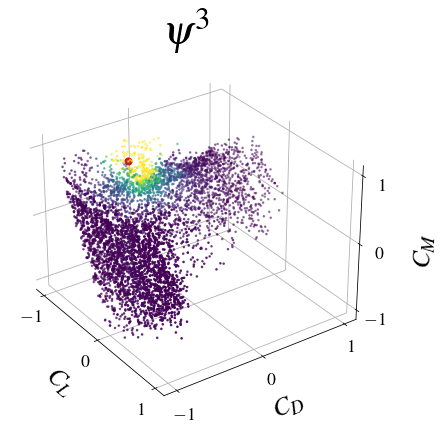}
  \caption{}
  \label{fig:airfoil_results_d}
\end{subfigure}
\caption{\label{fig:airfoil_results} Results for the airfoil problem. (a) Left: low-fidelity data with points closest to cluser centroids (in blue). Center: bi-fidelity data points. Right: high-fidelity data points used for validation. (b) Error distribution for the low- and bi-fidelity data. (c) The first three non-trivial eigenfunctions. (d) Three typical influence functions.}
\end{figure}

\paragraph{Results.} We sample $\Nl=27,000$ instances of input parameter vectors from a multivariate uniform distribution. 
Following Step 1, we employ XFOIL to generate the set of low-fidelity data. Then, we construct the graph Laplacian (with self-tuning $\sigma$, and $p,\,q=0.5$) and compute its eigendecomposition. The first three non-trivial eigenfunctions are shown in Figure \ref{fig:airfoil_results_c} in the normalized 3-dimensional data space.

For Step 2, we embed the low-fidelity data points in the eigenfunction space, and use K-means clustering to find $\Nh=70$ clusters and locate points closest to their centroids. 
Thereafter, we run CFD simulations to acquire the high-fidelity data at these points.
In the leftmost plots in Figure \ref{fig:airfoil_results_a}, we have plotted the low fidelity data projected onto the three independent planes, along with the points closest to the clusters centroids (shown in blue). We run $N_{val}=450$ additional high-fidelity simulations, corresponding to randomly selected low-fidelity data points, and use this validation set to compare the performance of the low- and bi-fidelity data. These validation data points are shown in the rightmost plots in Figure \ref{fig:airfoil_results_a}.

 To determine the transformation (\ref{eq:mftransf}), the minimization problem (\ref{eq:defpi}) is solved  with a spectrum cutoff of $K=210$, a value of $\tau=9.07 \cdot 10^{-4}$, and a regularization parameter $\omega^*=1.13 \cdot 10^{-6}$. A sample of three influence functions plotted over the low-fidelity dataset is shown in Figure \ref{fig:airfoil_results_d}. The resulting bi-fidelity data points are plotted in Figure \ref{fig:airfoil_results_a}. When compared with the low-fidelity data, we observe a significant upward shift in the moment coefficient for the bi-fidelity data. We also observe a compression of the data point cloud in the lift versus drag plane.
Overall, the bi-fidelity data distribution more closely resembles the distribution of the high-fidelity validation data. 

The average error for the the low- and bi-fidelity models, as defined in (\ref{eq:error}), is reported in Table \ref{tab:errors}, and their distribution is plotted in Figure \ref{fig:airfoil_results_d}.
We observe that the average error for lift an drag coefficients drops by a factor greater than two, while for the pitching moment it drops more than four times.
We also observe that the histograms of the distributions of the error for the bi-fidelity data are centered closer to zero and show a narrower spread. 

\section{Conclusions} 
\label{conc}

We have proposed a bi-fidelity approach to study the response of a system when two computational models of different fidelity and cost are available. The method follows three steps: \emph{(i)} acquire a large number of low-fidelity data, \emph{(ii)} identify and acquire a small number of key high-fidelity data, and \emph{(iii)} use the high-fidelity data to improve the accuracy of all low-fidelity data. 
This is accomplished by constructing an undirected, complete graph from the low-fidelity data points and computing its graph Laplacian. The low-lying spectrum of the graph Lalpacian is then used to cluster the low-fidelity data and to determine points closest to the centroids of the clusters. Thereafter, high-fidelity data is acquired only for these select points. This data, along with the spectral decomposition of the graph Laplacian, are used to solve a minimization problem which yields a transformation that maps each low-fidelity data point to new bi-fidelity coordinates. It is shown that this minimzation problem is convex. In numerical experiments, the approach yields bi-fidelity data that is significantly more accurate that its low-fidelity counterpart. In particular, in a problem motivated by biomechanics, this approach improves the accuracy of 4,000 low-fidelity data points by a factor which varies from 5-9 (depending on the quantity of interest) by only relying on 30 high-fidelity simulations. Similarly, in a problem of aerodynamics, it improves the accuracy of 27,000 low-fidelity data points by a factor of 2-4 while only using 70 high-fidelity simulations.

\section*{Acknowledgements}
This work was supported by ARO grant W911NF2010050 and the Army/Navy/NASAVertical Lift Research Center of Excellence (VLRCOE) Program, grant number W911W61120012, with Mahendra Bhagwat as technical monitor.

\bibliographystyle{unsrturl} 
\bibliography{references}  

\newpage
\appendix

\allowdisplaybreaks
\section{Gradient of the objective function}
\label{appx_gradient}

In this section, we evaluate an explicit expression for the gradient of the loss function $\mathsf{J}$ in (\ref{eq:defpi}) with respect to the parameters $\boldsymbol{\alpha}$. 

The gradient of the data misfit term is given by
\begin{align}
\frac{\partial\mathsf{J}_{\mathrm{data}}}{\partial\alpha_{uv}} & =\frac{2}{\Nh}\sum_{i=1}^{\Nh}\left(\mfdpi-\hfdpi\right) \cdot  \frac{\partial\mfdpi}{\partial\alpha_{uv}} \nonumber \\
 & =\frac{2}{\Nh}\sum_{i=1}^{\Nh}\left(\mfdpi-\hfdpi\right) \cdot \Big( \sum_{j=1}^{\Nh}\left(\hfdpj - \lfdpj \right)\frac{\partial\psi_{i}^{(j)}}{\partial\alpha_{uv}} \Big) \nonumber \\
 & =\frac{2}{\Nh}\sum_{i=1}^{\Nh}\sum_{j=1}^{\Nh}\left(\mfdpi-\hfdpi\right) \cdot \left(\hfdpj - \lfdpj\right)\phi_{i}^{(v)}\psi_{i}^{(j)}\left(\delta_{ju}-\psi_{i}^{(u)}\right). \label{eq:grad_data}
\end{align}
In deriving this expression we have  made use of
\begin{align}
\frac{\partial\psi_{i}^{(j)}}{\partial\alpha_{uv}} & =\frac{\partial}{\partial\alpha_{uv}}\frac{\exp{(v_{i}^{(j)})}}{\sum_{k}\exp{(v_{i}^{(k)})}} \nonumber \\
 & =\frac{\sum_{k}\exp{(v_{i}^{(k)})}\cdot \exp{(v_{i}^{(j)})}\phi_{i}^{(v)}\delta_{ju}-\exp{(v_{i}^{(j)})}\exp{(v_{i}^{(u)})}\phi_{i}^{(v)}}{\left[\sum_{k}\exp{(v_{i}^{(k)})}\right]^{2}} \nonumber \\
 & =\phi_{i}^{(v)}\psi_{i}^{(j)}\left(\delta_{ju}-\psi_{i}^{(u)}\right),
\end{align}
and 
\begin{align}
\frac{\partial v_{i}^{(j)}}{\partial\alpha_{uv}} & =\frac{\partial}{\partial\alpha_{uv}}\sum_{m=1}^{K}\alpha_{jm}\phi_{i}^{(m)}=\phi_{i}^{(v)}\delta_{ju}. \label{eq:grad_reg}
\end{align}
Here $\delta{}_{ij}$ is the Kronecker delta. 
The gradient of the regularization term is given by
\begin{align}
\frac{\partial\mathsf{J}_{\mathrm{reg}}}{\partial\alpha_{uv}} &= \frac{1}{K \Nh} \sum_{j=1}^{\Nh} \sum_{m=1}^{K} \frac{\partial}{\partial\alpha_{uv}}  \alpha_{jm}^2 \left(1 + \frac{\lambda_m}{\tau} \right)^2 \nonumber \\
 & =\frac{2\alpha_{uv}}{K \Nh}\left(1+\frac{\lambda_{v}}{\tau}\right)^2. 
\end{align}

Combining (\ref{eq:grad_data}) and (\ref{eq:grad_reg}) we have the expression for the total gradient,
\begin{equation}
\frac{\partial\mathsf{J}
}{\partial\alpha_{uv}} = \frac{2}{\Nh}\sum_{i=1}^{\Nh}\sum_{j=1}^{\Nh}\left(\mfdpi-\hfdpi\right) \cdot \left(\hfdpj - \lfdpj\right)\phi_{i}^{(v)}\psi_{i}^{(j)}\left(\delta_{ju}-\psi_{i}^{(u)}\right) + \frac{2 \omega \alpha_{uv}}{K \Nh}\left(1+\frac{\lambda_{v}}{\tau}\right)^2. \label{eq:grad_tot} 
\end{equation}

\section{Positive-definiteness of the Hessian of the objective function}
\label{appx_hessian}

The Hessian can be obtained by differentiating the gradient in (\ref{eq:grad_data}) and (\ref{eq:grad_reg}) with respect to the optimization parameters. This yields,

\begin{align}
H_{pquv} & \equiv \frac{\partial^2 \mathsf{J}_{\mathrm{data}}}{\partial\alpha_{pq} \partial\alpha_{uv}} \nonumber \\
 & =\frac{2}{\Nh} \Big[ \sum_{i=1}^{\Nh} ||\mfdpi-\hfdpi||  \frac{\partial^2 z^i}{\partial\alpha_{pq} \partial\alpha_{uv}} +  \sum_{i=1}^{\Nh} \frac{\partial \mfdpi}{ \partial\alpha_{pq}}  \cdot  \frac{\partial \mfdpi}{ \partial\alpha_{uv}} + \frac{\omega }{K}\big(1+\frac{\lambda_{v}}{\tau}\big)^2 \delta_{pu} \delta_{qv} \Big] ,
 \label{eq:hessian_data}
\end{align}

where $z^i \equiv \frac{\mfdpi-\hfdpi}{||\mfdpi-\hfdpi||} \cdot \mfdpi$. 
In the equation above, the first two terms emerge from the data-misfit term, while the third term is the contribution of the regularization term. The inner product of the Hessian with an arbitrary realization of parameters denoted by $\tilde{\bf{\alpha}}$ writes

\begin{align}
\sum_{pquv} \tilde{\alpha}_{pq} H_{pquv} \tilde{\alpha}_{uv} 
 & =\frac{2}{\Nh} \Big[ \sum_{i=1}^{\Nh} ||\mfdpi-\hfdpi|| \sum_{pquv} \tilde{\alpha}_{pq} 
 \frac{\partial^2 z^i}{\partial\alpha_{pq} \partial\alpha_{uv}} \tilde{\alpha}_{uv} +  \sum_{i=1}^{\Nh} \big(\sum_{pq}  \frac{\partial \mfdpi}{ \partial\alpha_{pq}} \tilde{\alpha}_{pq}\big)^2 + \frac{\omega }{K} \sum_{pq} \big(1+\frac{\lambda_{q}}{\tau}\big)^2 \tilde{\alpha}_{pq}^2 \Big].
 \label{eq:hessian_data2}
\end{align}

In the equation above, the term $\frac{\partial^2 z^i}{\partial\alpha_{pq} \partial\alpha_{uv}}$ is a symmetric fourth order tensor and therefore has real eigenvalues. We let $\zeta =  \lambda_{\rm min}(\frac{\partial^2 z^i}{\partial\alpha_{pq} \partial\alpha_{uv}} ),i = 1,\dots, \Nh$ be the smallest eigenvalue. Therefore, we have 

\begin{align}
\sum_{pquv} \tilde{\alpha}_{pq} H_{pquv} \tilde{\alpha}_{uv} 
 & \ge \frac{2}{\Nh} \Big[ \zeta \sum_{i=1}^{\Nh} ||\mfdpi-\hfdpi||  \sum_{pq} \tilde{\alpha}_{pq}^2  +  \sum_{i=1}^{\Nh} \big( \sum_{pq}  \frac{\partial \mfdpi}{ \partial\alpha_{pq}} \tilde{\alpha}_{pq} \big)^2 + \frac{\omega }{K} \sum_{pq} \tilde{\alpha}_{pq}^2 \Big] ,
 \label{eq:hessian_ip}
\end{align}

With $\zeta \ge 0$, without any restriction on the data misfit, we have 
\begin{align}
\sum_{pquv} \tilde{\alpha}_{pq} H_{pquv} \tilde{\alpha}_{uv} 
 & \ge \frac{2 \omega }{\Nh K} \sum_{pq} \tilde{\alpha}_{pq}^2,
\end{align}
which implies that the Hessain is positive definite. 
With $\zeta < 0$, we require $ \sum_{i=1}^{\Nh} ||\mfdpi-\hfdpi|| \le \frac{\omega }{ 2 |\zeta| K }$, and use this in (\ref{eq:hessian_ip}) to arrive at 
\begin{align}
\sum_{pquv} \tilde{\alpha}_{pq} H_{pquv} \tilde{\alpha}_{uv} 
 & \ge \frac{\omega }{\Nh K} \sum_{pq} \tilde{\alpha}_{pq}^2,
\end{align}

Thus, assuming $ \sum_{i=1}^{\Nh} ||\mfdpi-\hfdpi|| \le \frac{\omega }{ 2 |\zeta| K }$ we conclude that the Hessian is positive definite. 

\section{Solution for a canonical problem}
\label{appx_canonical}

We make the following assumptions on the data:
\begin{enumerate}
    \item The low-fidelity data $\lfdpi$ are distributed among $M$ distinct clusters.
    \item  $C(i)$ is the cluster number for the $i-$th data point. Without loss of generality, we may assume that points are numbered such that $C(i) = i$ for $i = 1, \dots, M$.
    \item For any point $i$, the high-fidelity data is related to the low-fidelity data via the transformation
    \begin{equation}
        \hfdpi = \lfdpi + \bm{\delta}^{C(i)}.
    \end{equation}
    where $\bm{\delta}^{j}, j = 1, \dots, M$, are the rigid translations for each cluster. 
    \item For $i = 1, \dots, M$, the high-fidelity data $\hfdpi$ is available and is used to construct the data matching term. Using assumptions 2 and 3, we note that these points,
        \begin{equation}
        \hfdpi = \lfdpi + \bm{\delta}^{i}, \qquad i = 1, \dots, M. \label{eq:hf_spec2}
    \end{equation}
\end{enumerate} 
Under the assumptions above, the proposed method admits a bi-fidelity solution whose error vanishes as the regularization parameter  approaches zero. 

\begin{figure}
    \centering
    \includegraphics[width=0.65\textwidth]{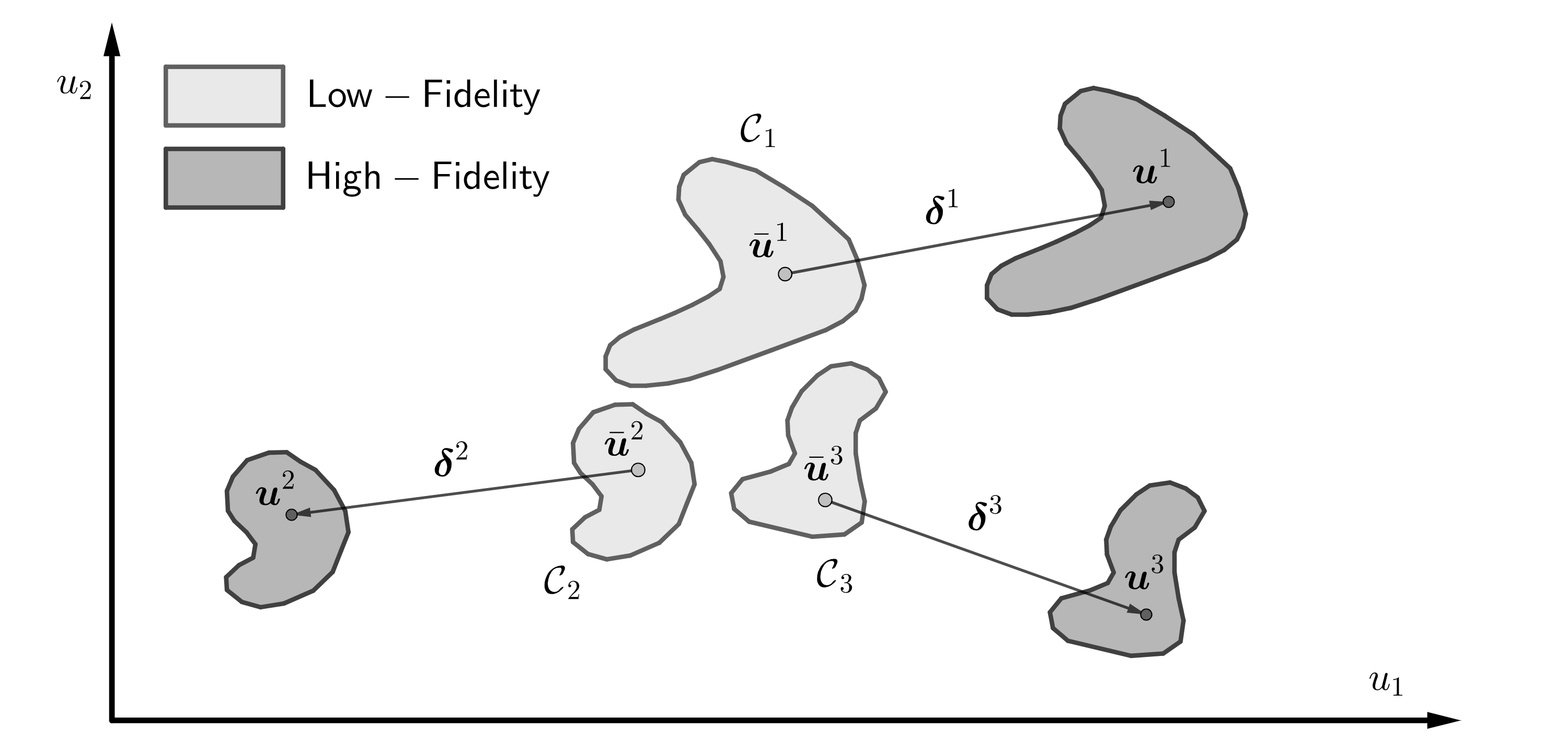}
    \caption{Schematic of a canonical problem with the dataset partitioned into $M=3$ clusters.}
    \label{fig:canonical_problem}
\end{figure}

\paragraph{Proof.} In the limit of infinite data, the graph Laplacian of the low-fidelity graph yields $M$ eigenfunctions with null eigenvalues \cite{hoffmann2020consistency} such that,
\begin{equation}
    \phi_i^{(j)} = \left\{ \begin{array}{rl}  1, & C(i) = j \\ -1, & C(i) \ne j \end{array} \right. . \label{eq:efun_spec}
\end{equation}
That is, the $j$-th eigenfunction attains a value  of $+ 1$ for all points that belong to the $j$-th cluster and a value of $-1$ for all other points. 

For the proposed method we consider the choice $K = \hat{N} = M$. Further, we consider the special case where the unknown parameters are given by, $\alpha_{jm} = \alpha \delta_{jm}$. With this choice, from (\ref{eq:efun_spec}) and (\ref{eq:defv}) we conclude that 
\begin{equation}
    v_i^{(j)} = \left\{ \begin{array}{rl}  \alpha, & C(i) = j \\ - \alpha, & C(i) \ne j \end{array} \right. . \label{eq:v_spec}
\end{equation}
From (\ref{eq:defpsi}) this yields,
\begin{equation}
    \psi_i^{(j)} = \left\{ \begin{array}{rl}  \frac{1}{1+ (M-1) \exp(-2 \alpha)}, & C(i) = j \\ \frac{\exp(-2 \alpha)}{1+ (M-1) \exp(-2 \alpha)}, & C(i) \ne j \end{array} \right. . \label{eq:psi_spec}
\end{equation}

For $i = 1, \dots, M$, the difference between the the multi-fidelity expression (\ref{eq:mftransf}) and the high-fidelity data (\ref{eq:hf_spec2}) is given by 
\begin{eqnarray*}
\mfdpi  - \hfdpi &=& \lfdpi + \sum_{j = 1}^{M} \bm{\delta}_j \psi^{(j)}_i - \big(\lfdpi + \bm{\delta}^{i}\big)  \nonumber \\
    &=& \bm{\delta}^{i} (\psi^{(i)}_i  - 1) + \sum_{j = 1, j \ne i }^{M} \bm{\delta}^j \psi^{(j)}_i \nonumber \\
    &=& \frac{\exp(-2 \alpha)}{1+ (M-1) \exp(-2 \alpha)} \overbrace{\big( -(M-1) \bm{\delta}^{i} +  \sum_{j = 1, j \ne i }^{M} \bm{\delta}^j \big)}^{ \equiv \bm{b}^i} , \qquad \mbox{from (\ref{eq:psi_spec}})\nonumber \\
    &=& \gamma (\alpha) \bm{b}^i.  
\end{eqnarray*}
Where $\gamma (\alpha) \equiv \frac{\exp(-2 \alpha)}{1+ (M-1) \exp(-2 \alpha)}$. 
Using this expression in the definition of the data matching term (\ref{eq:piterm-data}), we arrive at,
\begin{equation}
     \mathsf{J}_{\mathrm{data}}(\alpha) =  \gamma (\alpha)^2 \sum_{i = 1}^{M}  \frac{||\bm{b}^i||^2}{M}. 
\end{equation}

Using (\ref{eq:v_spec}) in (\ref{eq:piterm-reg}) we conclude that the regularization term is given by, 
\begin{equation}
\mathsf{J}_{\mathrm{reg}}(\alpha) =  \frac{\alpha^2}{M},
\end{equation}
where we have used the fact the eigenvalues for the eigenfunctions considered in this expansion are zero. 

Therefore the total objective function is equal to,
\begin{equation}
     \mathsf{J}(\alpha) =\gamma(\alpha)^2  \sum_{i = 1}^{M}  \frac{||\bm{b}^i||^2}{M} + \omega  \frac{\alpha^2}{M},
\end{equation}
Further, it is easily verified that 
\begin{equation}
    \frac{d \gamma}{d \alpha } = 2 \gamma \big[ (M-1) \gamma - 1 \big].
\end{equation}
Setting $\frac{d \mathsf{J} }{d \alpha} = 0$ to find the stationary point of the objective function, we arrive at, 
\begin{equation}
    2 \gamma \big[ (M-1) \gamma - 1 \big] \sum_{i = 1}^{M}  \frac{||\bm{b}_i||^2}{M} + \omega  \frac{\alpha}{M} = 0.
\end{equation}
In the limit $\omega \to 0$, to leading order this equation yields the solution 
\begin{equation}
    \alpha = - \frac{\ln \omega}{4}. 
\end{equation}

Using this solution in the expression of the data matching term we note that $\mathsf{J}_{\mathrm{data}}(\alpha)  = O(\omega)$, which tends to zero as $\omega \to 0$. Also $\mathsf{J}_{\mathrm{reg}}(\alpha) = O( \omega^2 \ln^2 \omega)$ which also tends to zero as $\omega \to 0$. Thus both the data matching and regularization terms tend to zero as the regularization parameter tends to zero. 

We now show that with choice of $\alpha$, for any point indexed by $i$, the coordinates predicted by the multi-fidelity approximation tend to the high-fidelity coordinates as the regularization is reduced. To accomplish this we compute 

\begin{eqnarray*}
\mfdpi  - \hfdpi &=& \lfdpi + \sum_{j = 1}^{M} \bm{\delta}^j \psi^{(j)}_i - \big(\lfdpi + \bm{\delta}^{C(i)}\big)  \nonumber \\
    &=& \bm{\delta}^{C(i)} (\psi^{(C(i))}_i  - 1) + \sum_{j = 1, j \ne C(i) }^{M} \bm{\delta}^j \psi^{(j)}_i \nonumber \\
    &=& \frac{\exp(-2 \alpha)}{1+ (M-1) \exp(-2 \alpha)} \big[ -(M-1) \bm{\delta}^{C(i)} +  \sum_{j = 1, j \ne C(i) }^{M} \bm{\delta}^j \big], \qquad \mbox{from (\ref{eq:psi_spec}})\nonumber \\
    &=& \gamma (\alpha) \bm{b}^{C(i)} .  
\end{eqnarray*}

Therefore, 

\begin{equation}
    ||\mfdpi  - \hfdpi|| = \gamma (\alpha) || \bm{b}^{C(i)}|| = || \bm{b}^{C(i)}|| \omega^{1/2}( 1 + O(\omega)). 
\end{equation}

Clearly this difference tends to zero as $\omega \to 0$. 

\section{L-curve criterion for determining the regularization parameter} \label{appx_Lcurve}

\begin{figure}
    \centering
    \includegraphics[width=0.45\textwidth]{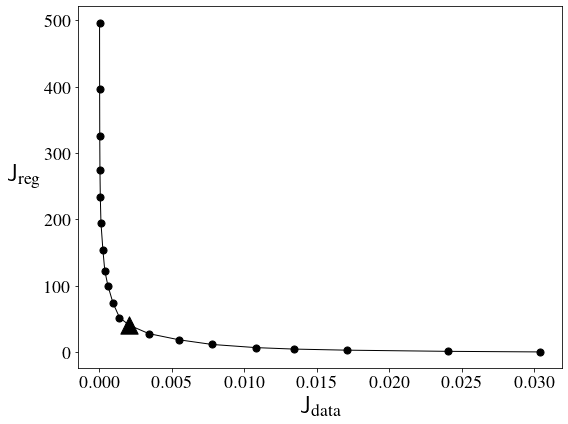}
    \caption{L-curve for the traction on soft material problem of Section \ref{sec:tracion_prob}. The optimal value for the regularization parameter is the one corresponding to the elbow of the curve, marked with a triangle in the graph.}
    \label{fig:l-curve}
\end{figure}

The L-curve is a plot of the data misfit loss $\mathsf{J}_{\mathrm{data}}(\bm{\alpha}^*(\omega); \, \omega)$ versus the regularization loss $\mathsf{J}_{\mathrm{reg}}(\bm{\alpha}^*(\omega); \, \omega)$ obtained after solving the minimization problem (\ref{eq:defpi}) for different values of regularization parameter $\omega$. It can be used to visualize the balance between the two terms and provides a way to tune the regularization parameter. The optimal value $\omega^*$ is indeed chosen to be the one  corresponding to the elbow of the L-curve, i.e. the point of maximum curvature. In fact, the elbow of the curve separates the regions where the final solution is either dominated by a large data misfit error or by a large regularization loss. Therefore, any perturbation of the parameter $\omega^*$ would lead to a significant increase in one loss term or the other. 
Because the minimization problem (\ref{eq:defpi}) is inexpensive to solve, we can use the L-curve criterion to determine the optimal value of $\omega$. In Figure \ref{fig:l-curve} we show the L-curve used to determine the regularization parameter for the traction on soft material problem in Section \ref{sec:tracion_prob}. 

\section{Effect of the selection strategy of high-fidelity data}
In Step 2 of the method, we propose a strategy to select the parameters values at which to acquire high-fidelity data.
The strategy consists in determining the low-fidelity data points that are the closest to the centroids of the clusters arising within the low-fidelity data, and then employ the high-fidelity model to compute their counterpart.

We want to compare the performance of this strategy against a random selection of the low-fidelity points for which we acquire a high-fidelity version. 
To do that, we consider the problem of predicting the traction on a soft body and the results obtained in Section \ref{sec:tracion_prob}. 
Then, we solve the same problem with a random selection of the parameters at which to compute high-fidelity data points. 
The number of high-fidelity data points used is fixed, $\Nh=30$, and the problem is solved 200 times with different random selections. Then, we compute the mean and standard deviation of the error for each component as defined in (\ref{eq:error}), and denote them as $M_m$ and $\Sigma_m$, $m=1,\,\dots,\,D$, respectively. The results are reported in Table \ref{tab:errors_selection_strategy}. 

We notice that the errors committed with a random selection is larger than the errors attained with the proposed selection approach, but is still considerably smaller than the low-fidelity errors. 
This suggests that the proposed selection strategy ought to be preferred to a random choice when possible, but that the bi-fidelity method can be successfully applied even when the set of high-fidelity data is predetermined.   

\begin{table}[hbt!]
\centering
\begin{tabular}{|c|c|c c c c c|}
\hline 
\multicolumn{2}{|c|}{\textbf{Error {[}\%{]}}} & \multicolumn{5}{c|}{\textbf{Soft body with inclusion}}\tabularnewline
\hline 
\multicolumn{2}{|c|}{\rule{0pt}{3ex} Quantity of interest } & $t_{1}$ & $t_{2}$ & $t_{3}$ & $t_{4}$ & $\sigma_{yy}^{\mathrm{max}}$\tabularnewline
\hline 
\multicolumn{2}{|c|}{\rule{0pt}{3ex} Low-fidelity} & 4.48 & 7.15 & 7.21 & 4.65 & 10.19\tabularnewline
\hline 
\multicolumn{2}{|c|}{\rule{0pt}{3ex} Bi-fidelity with proposed selection strategy} & \textbf{0.50} & \textbf{1.37} & \textbf{1.39} & \textbf{0.67} & \textbf{2.20}\tabularnewline
\hline 
\multirow{2}{*}{\rule{0pt}{3ex} Bi-fidelity with random selection} & \rule{0pt}{3ex} $M_m$ & 0.68 & 1.54 & 1.50 & 0.81 & 2.42\tabularnewline
\cline{2-7} \cline{3-7} \cline{4-7} \cline{5-7} \cline{6-7} \cline{7-7} 
 & \rule{0pt}{3ex} $\Sigma_m$ & 0.11 & 0.16 & 0.16 & 0.12 & 0.18\tabularnewline
\hline 
\end{tabular}
\vspace{2mm}
\caption{\label{tab:errors_selection_strategy} Error of the bi-fidelity model constructed with the proposed selection strategy and a random selection.}
\end{table}

\section{Traction on soft material with a stiff inclusion}
\label{appx_traction}
\begin{figure}
\centering
\includegraphics[width=.35\textwidth]{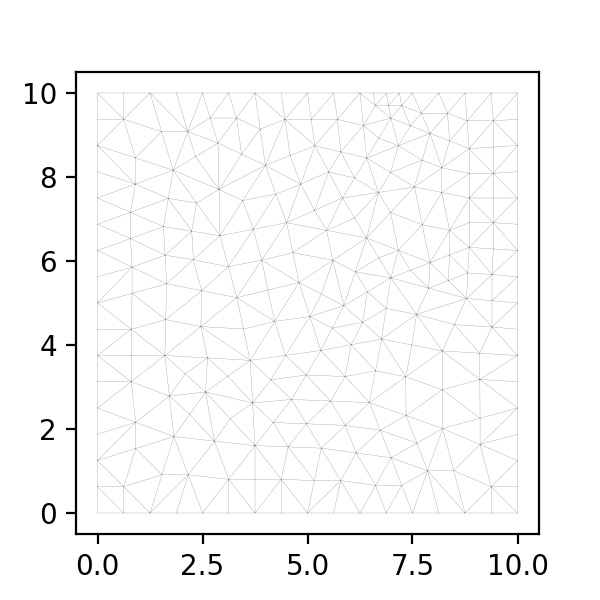}
\includegraphics[width=.35\textwidth]{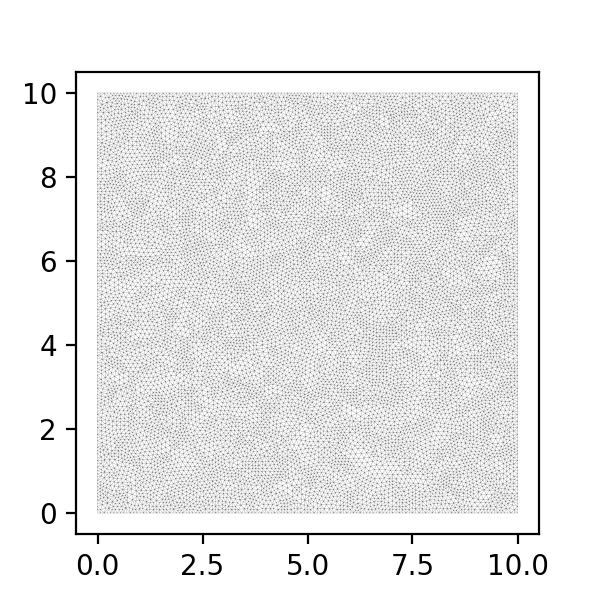}
\caption{Comparison between the two meshes of the low- and high-fidelity finite element models used to solve the traction on soft material problem.}
\label{FEmesh}
\end{figure}

The low- and high-fidelity models used to solve the elasticity problem described in Section \ref{sec:tracion_prob} are finite element solvers that differ by their mesh density, i.e. the number of elements used to discretize the computational domain. In Figure \ref{FEmesh} we show a comparison between the two meshes. 

\begin{figure}
    \centering
    \includegraphics[width=0.9\textwidth]{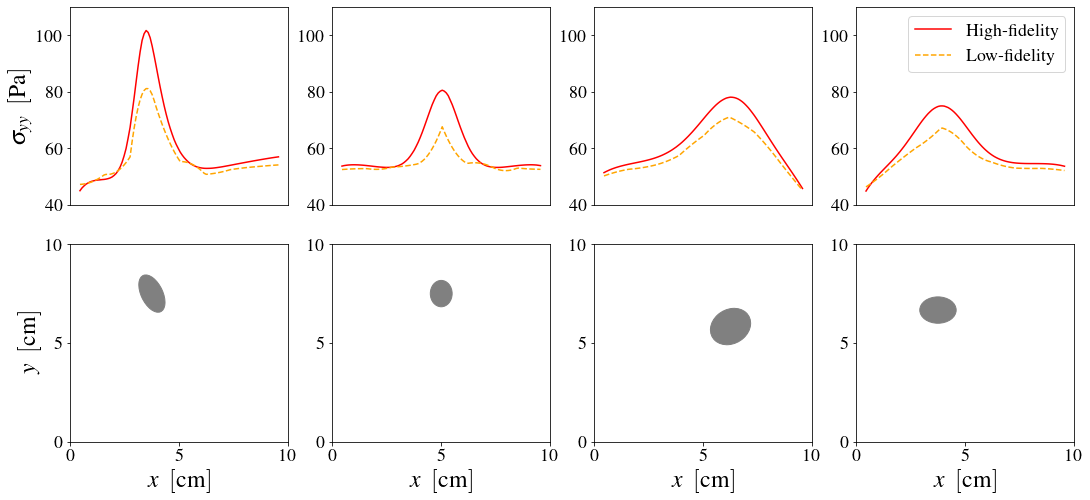}
    \caption{Sample of low- and high-fidelity traction fields for different inclusions.}
    \label{fig:tract_fields}
\end{figure}

The low- and high-fidelity solutions of the problem show similar trends, but the low-fidelity solutions tend to underestimate the magnitude of traction field. In Figure \ref{fig:tract_fields} we show a sample of low- and high-fidelity solutions for different instances of input parameters. 

\section{Aerodynamic coefficients of NACA 0012}
\label{appx_airfoil}
In Section \ref{sec:airfoil} we consider the problem of predicting the aerodynamic coefficients of 4-digit NACA airfoils for different Reynolds number and angles of attack. After applying the proposed method,
we show the resulting bi-fidelity dataset in Figure \ref{fig:airfoil_results_a} as a point cloud in the 3-dimensional data space of the lift, drag and moment coefficients ($C_L, \, , C_D$ and $C_M$, respectively).

\begin{figure}
    \centering
    \includegraphics[width=0.9\textwidth]{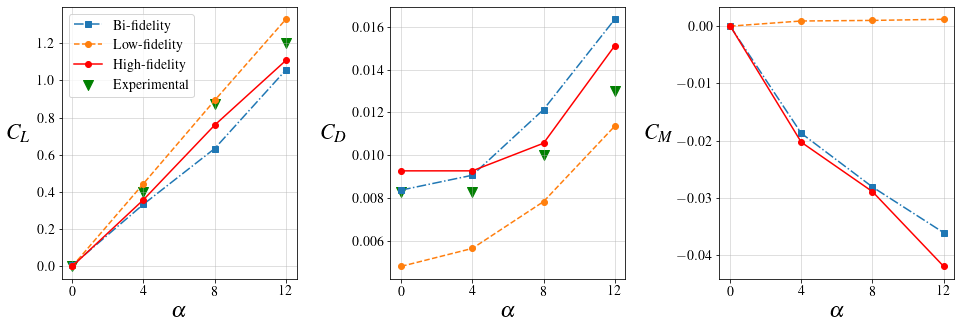}
    \caption{Comparison of the curves of the lift, drag and moment coefficients versus angle of attack for the airfoil NACA 0012 at $Re=6\cdot10^6$. The low-fidelity data are indicated with a dashed orange line, the bi-fidelity data with a dash-dot blue line, the high-fidelity data with a solid red line, and the experimental data are marked with a green upside-down triangle.}
    \label{fig:polar_comparison}
\end{figure}

To better visualize the data at the three levels of fidelity, we select the case of the NACA 0012 airfoil profile at Reynolds $Re = 6 \cdot 10^6$, and plot, in Figure \ref{fig:polar_comparison}, the graphs of the aerodynamic coefficients versus angle of attack for the three models, including experimental data available for lift and drag coefficients \cite{Ladson1988}.

\begin{figure}
    \centering
    \includegraphics[width=0.9\textwidth]{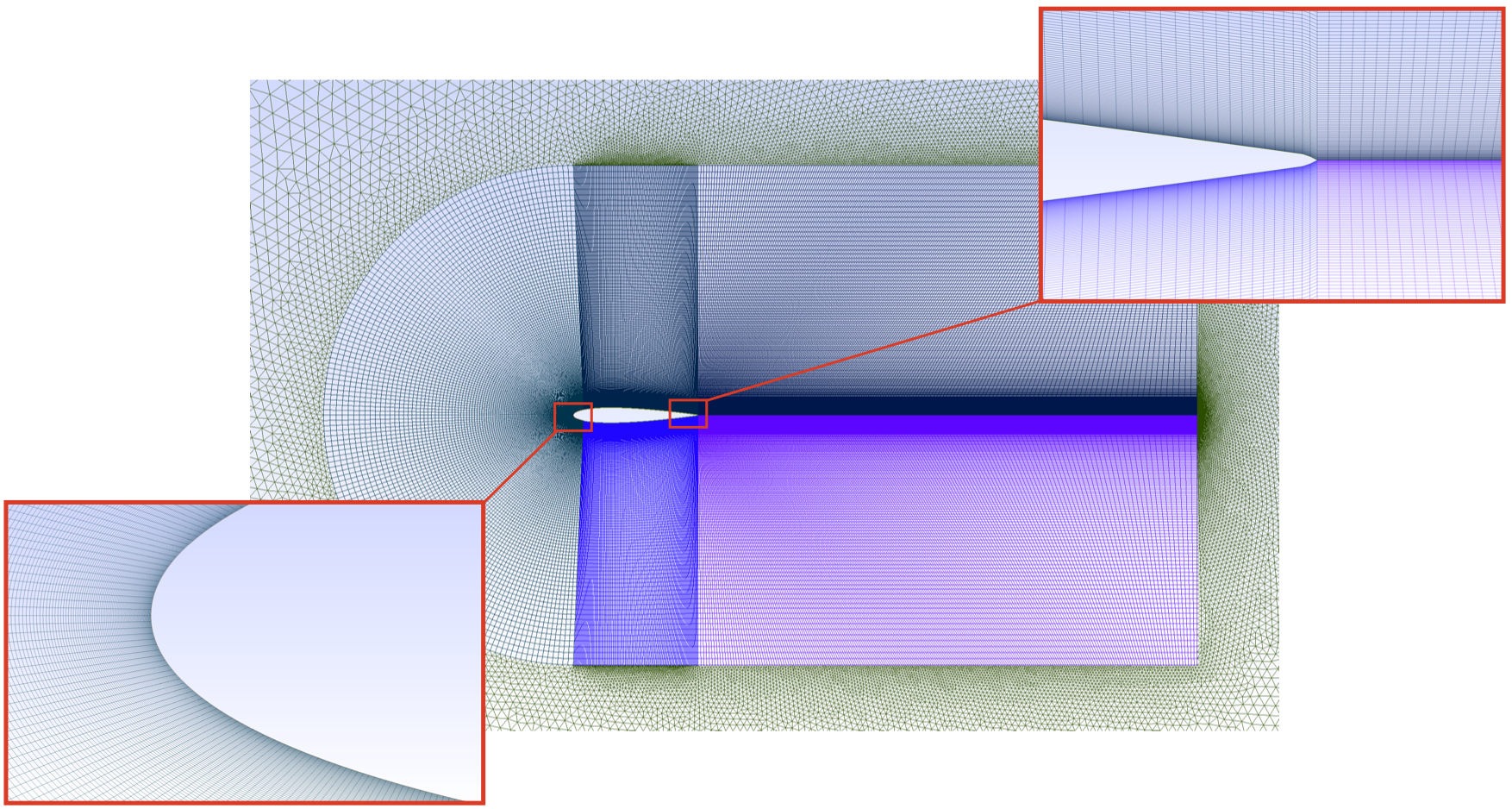}
    \caption{Example of a mesh used for the high-fidelity CFD simulations.}
    \label{fig:cfd_mesh}
\end{figure}

We observe that the high-fidelity CFD simulations match the experimental results within 10\% error for lift and 15\% error for drag. In Figure \ref{fig:cfd_mesh} we show the computational mesh used to perform these simulations.

By analyzing the lift and drag coefficients curves, we note that the bi-fidelity model could learn and retain the low-fidelity trend, and adjust the magnitudes in light of the sparse high-fidelity data. 
For the moment coefficient, the low-fidelity data show a significant disagreement with the high-fidelity results. Nonetheless, the predictions of the bi-fidelity model closely match the high-fidelity data both in trend and magnitude. 
We conclude that the bi-fidelity model could correct the data structures in the 3-dimensional data space using the few select high-fidelity data points available, and has increased the accuracy for all low-fidelity points.

\section{Definition of the data space} \label{appx_dataspace} 

As discussed in Section \ref{bi-fi-section}, we consider problems where the solution for a given instance of input parameters $\pari$ is a set of quantities of interest $\qoi(\pari)$. 

The data space $\dpt$ where the low-fidelity graph is constructed is defined as
$\dpt(\pari) = \bm{R}[\pari, \qoi (\pari)]$, with $\bm{R}$ being a restriction operator that extracts the appropriate components of $\pari$ and $\qoi$.
The choice of $\bm{R}$ is problem dependent. It should include all the variables directly related to the predictions one wants to make, i.e. the subset of the relevant components of $\qoi$, and the parameters that would help identifying clusters and structures in the data. 

For example, including bifurcation parameters in the data space is important, as small changes in their value can cause significant changes in the topology or qualitative nature of the physical solution. Similarly, parameters that identify different and distinct regimes of the solution, e.g. the Reynolds number in fluid dynamics, provide valuable information about different data points, and can help separating the point cloud. This is especially important when the performance of the low-fidelity model strongly depend upon certain parameters. If the error of the low-fidelity data significantly differs based on a parameter, adding it to the data space will make sure that the transformation (\ref{eq:mftransf}) will act accordingly. 

\begin{figure}
    \centering
    \includegraphics[width=0.6\textwidth]{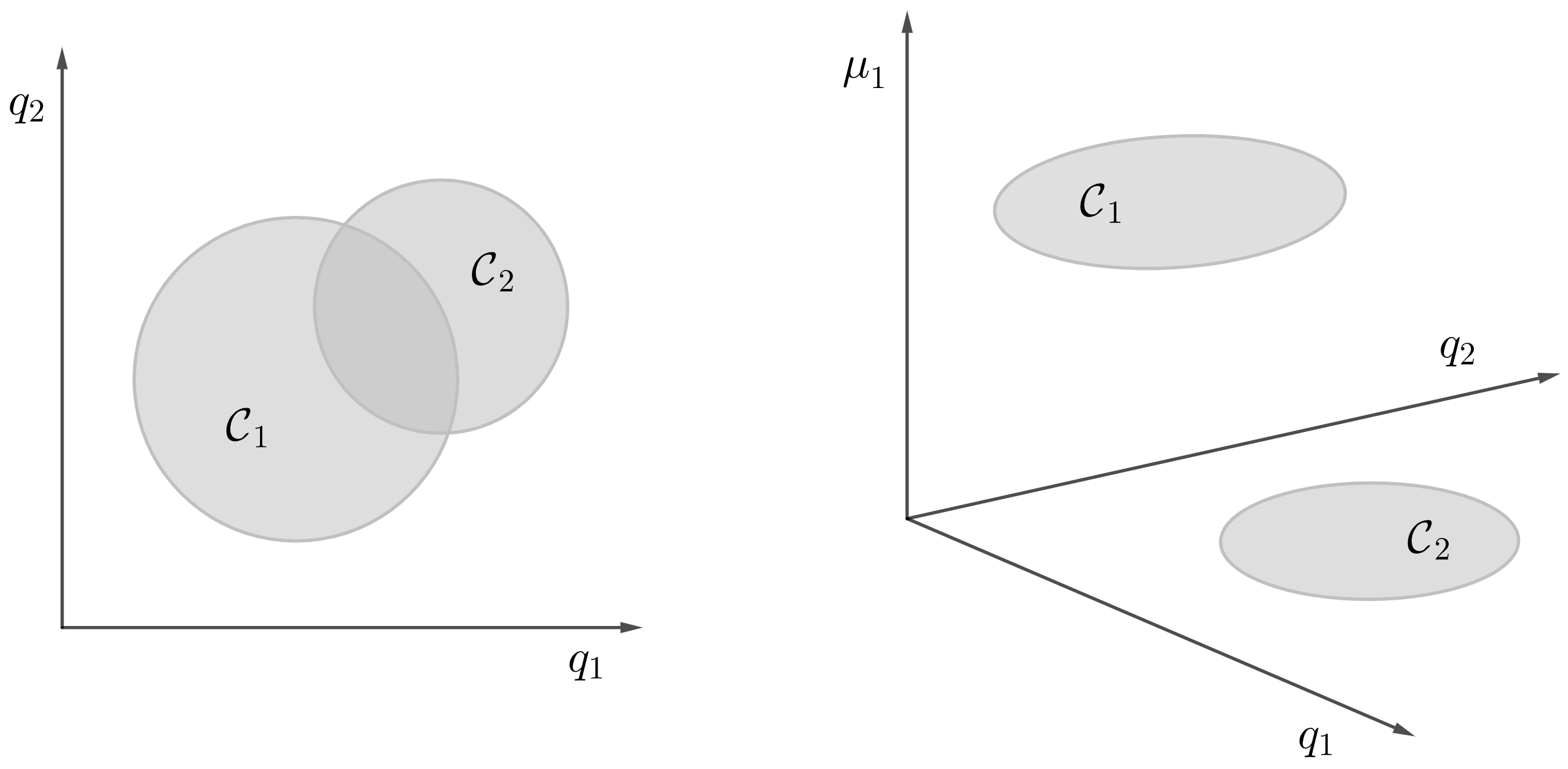}
    \caption{Example of a case where adding the parameter $\mu_1$ to the data space $(q_1,\, q_2)$ separates the data set into two distinct clusters, $\mathcal{C}_1$ and $\mathcal{C}_2$.}
    \label{fig:data-space}
\end{figure}

In Figure \ref{fig:data-space} we show a schematic of a simple case where adding the parameter $\mu_1$ to the data space $(q_1,\, q_2)$ leads to a clear separation of the data points. In this illustrative example, including the parameter in the data space allows for a better selection of the high-fidelity data to acquire and a more appropriate treatment of the two clusters.

It is important to notice that the transformation (\ref{eq:mftransf}) does not lead to any displacements along the parameters directions, as the high-fidelity data points will have by construction the same input parameters. Hence, the bi-fidelity transformation will act only on the space of the quantities of interests, as it should.

\subsection{Numerical results with augmented data space} 
 In this Section we analyze the results of the numerical problems proposed in Section \ref{num-experiment}, when the input parameters are included in data space. 

For the problem of traction on soft body (Section \ref{sec:tracion_prob}), with input parameters $\pari=[x_c,\,y_c,\,\theta,\,a,\,b]$ and quantities of interest $\qoi=[f_1,\,f_2,\,f_3,\,f_4,\,\sigma_{yy}^{\mathrm{max}}]$, we consider the case where the data space is formed by concatenating all inputs and all outputs, i.e. $\dpt = [\pari,\, \qoi]$.

On the other hand, for the problem of the aerodynamic coefficients of NACA airfoils (Section \ref{sec:airfoil}), the input parameters and the set of quantities of interest are $\pari=[\eta,\, x_\eta,\, t,\, \alpha,\, Re]$ and $\qoi = [C_L, \, C_D, \, C_M]$, respectively. In this case, we include only the Reynolds number in the data space, i.e. $\dpt = [Re, C_L, \, C_D, \, C_M]$.

For both problems, we apply once again the proposed bi-fidelity method considering the graph constructed in the augmented data space $\dpt = \bm{R}[\pari,\, \qoi]$. The numerical results are reported in Table \ref{tab:errors-2}.
We note that in these particular numerical experiments, using only the quantities of interest in the definition of the data space (that is $\dpt = \qoi$) leads to better results in almost all cases. 

\begin{table}[hbt!]
\centering
\begin{tabular}{|c | c c c c c | c c c|}
\cline{1-9} 
\multicolumn{1}{|c|} {\textbf{Error [\%]}} & \multicolumn{5}{c|}{\textbf{Soft body with inclusion}} & \multicolumn{3}{c|}{\textbf{Airfoil}}\tabularnewline
\hline 
\rule{0pt}{3ex} Quantity of interest & $f_{1}$ & $f_{2}$ & $f_{3}$ & $f_{4}$ & $\sigma_{yy}^{\mathrm{max}}$ & $C_{L}$ & $C_{D}$ & $C_{M}$\tabularnewline
\hline 
\rule{0pt}{3ex} Low-fidelity & 4.48 & 7.15 & 7.21 & 4.65 & 10.19 & 42.79 & 28.81 & 216.11 \tabularnewline
\rule{0pt}{3ex} Bi-fidelity with $\dpt = \qoi$ & \textbf{0.50} & \textbf{1.35} & \textbf{1.37} & \textbf{0.63} & 2.13 & \textbf{19.90} & \textbf{11.71} & \textbf{46.38} \tabularnewline
\rule{0pt}{3ex} Bi-fidelity with $\dpt = \bm{R}[\pari,\, \qoi]$ & 0.69 & 1.58 & 1.66 & 0.68 & \textbf{1.87} & 21.01 & 12.04 & 51.62 \tabularnewline
\hline 
\end{tabular}
\vspace{2mm}
\caption{\label{tab:errors-2} Error of the low- and bi-fidelity model for each quantity of interest.}
\end{table}

\end{document}